%% file: template.tex
\documentclass{article}

\usepackage{arxiv}
\usepackage[utf8]{inputenc}
\usepackage[T1]{fontenc}
\usepackage{url}
\usepackage{booktabs}
\usepackage{amsfonts}
\usepackage{nicefrac}
\usepackage{microtype}
\usepackage{lipsum}
\usepackage{graphicx}
\usepackage{enumitem}
\usepackage{wrapfig}
\usepackage{multirow}
\usepackage{wasysym}
\usepackage{makecell}
\usepackage{pifont}
\usepackage{colortbl}
\usepackage{caption}
\usepackage[hidelinks]{hyperref}
\usepackage{tabularx}
\usepackage{tcolorbox}
\usepackage{orcidlink}
\usepackage{xcolor}
\usepackage{soul}
\usepackage{array}
\usepackage{calc}
\usepackage{float}
\usepackage{tikz}
\usetikzlibrary{shapes.misc}

\graphicspath{ {./images/} }

\definecolor{colorVU}{RGB}{220, 248, 220}
\definecolor{colorDD}{RGB}{255, 225, 195}
\definecolor{colorML}{RGB}{235, 248, 255}
\definecolor{colorQA}{RGB}{250, 235, 250}
\definecolor{colorMM}{RGB}{255, 255, 220}
\definecolor{colorAvg}{RGB}{255, 228, 230}

\newcommand{\hlVU}[1]{{\sethlcolor{colorVU}\hl{#1}}}
\newcommand{\hlDD}[1]{{\sethlcolor{colorDD}\hl{#1}}}
\newcommand{\hlML}[1]{{\sethlcolor{colorML}\hl{#1}}}
\newcommand{\hlQA}[1]{{\sethlcolor{colorQA}\hl{#1}}}
\newcommand{\hlMM}[1]{{\sethlcolor{colorMM}\hl{#1}}}

\begin{document}

\begin{center}
\rule{\textwidth}{2.8pt}  

\vspace{1.1em}

{\Large\bfseries \textit{MedMASLab}: A Unified Orchestration Framework for Benchmarking Multimodal Medical Multi-Agent Systems}

\vspace{1.1em}
\rule{\textwidth}{1.0pt}  

\vspace{1.2em}  

{\normalsize  
\textbf{Yunhang Qian}$^{1\dagger}$ \quad
\textbf{Xiaobin Hu}$^{1\dagger}$ \quad
\textbf{Jiaquan Yu}$^{1,2\dagger}$ \quad
\textbf{Siyang Xin}$^{1,3\dagger}$  \quad
\textbf{Xiaokun Chen}$^{6}$ \\
\textbf{Jiangning Zhang}$^{4}$ \quad
\textbf{Peng-Tao Jiang}$^{5}$ \quad
\textbf{Jiawei Liu}$^{1,2}$ \quad
\textbf{Hongwei Bran Li}$^{1}$ \quad
}

\vspace{0.8em}

{\small
$^{1}$National University of Singapore \quad
$^{2}$University of Science and Technology of China \\
$^{3}$Fudan University \qquad
$^{4}$Zhejiang University \quad
$^{5}$vivo \quad
$^{6}$Stanford University \\
$^{\dagger}$Equal contribution
}

\end{center}

\vspace{1.5em}
\vspace{1.5em}
\begin{abstract}
While Multi-Agent Systems (MAS) show potential for complex clinical decision support, the field remains hindered by architectural fragmentation and the lack of standardized multimodal integration. Current medical MAS research suffers from non-uniform data ingestion pipelines, inconsistent visual-reasoning evaluation, and a lack of cross-specialty benchmarking. To address these challenges, we present \textbf{MedMASLab}, a unified framework and benchmarking platform for multimodal medical multi-agent systems. 
MedMASLab introduces: (1) A standardized multimodal agent communication protocol that enables seamless integration of 11 heterogeneous MAS architectures across 24 medical modalities. (2) An automated clinical reasoning evaluator, a zero-shot semantic evaluation paradigm that overcomes the limitations of lexical string-matching by leveraging large vision-language models to verify diagnostic logic and visual grounding. (3) The most extensive benchmark to date, spanning 11 organ systems and 473  diseases, standardizing data from 11 clinical benchmarks. Our systematic evaluation reveals a critical domain-specific performance gap: while MAS improves reasoning depth, current architectures exhibit significant fragility when transitioning between specialized medical sub-domains. We provide a rigorous ablation of interaction mechanisms and cost-performance trade-offs, establishing a new technical baseline for future autonomous clinical systems. The source code and data is publicly available at: \href{https://github.com/NUS-Project/MedMASLab/}{https://github.com/NUS-Project/MedMASLab/}

\keywords{Medical Multi-Agent Systems \and Multimodal Benchmarking \and Vision-Language Models \and Clinical Reasoning Verification}
\end{abstract}

\begin{figure}[t]
    \centering
    \includegraphics[width=0.9\textwidth]{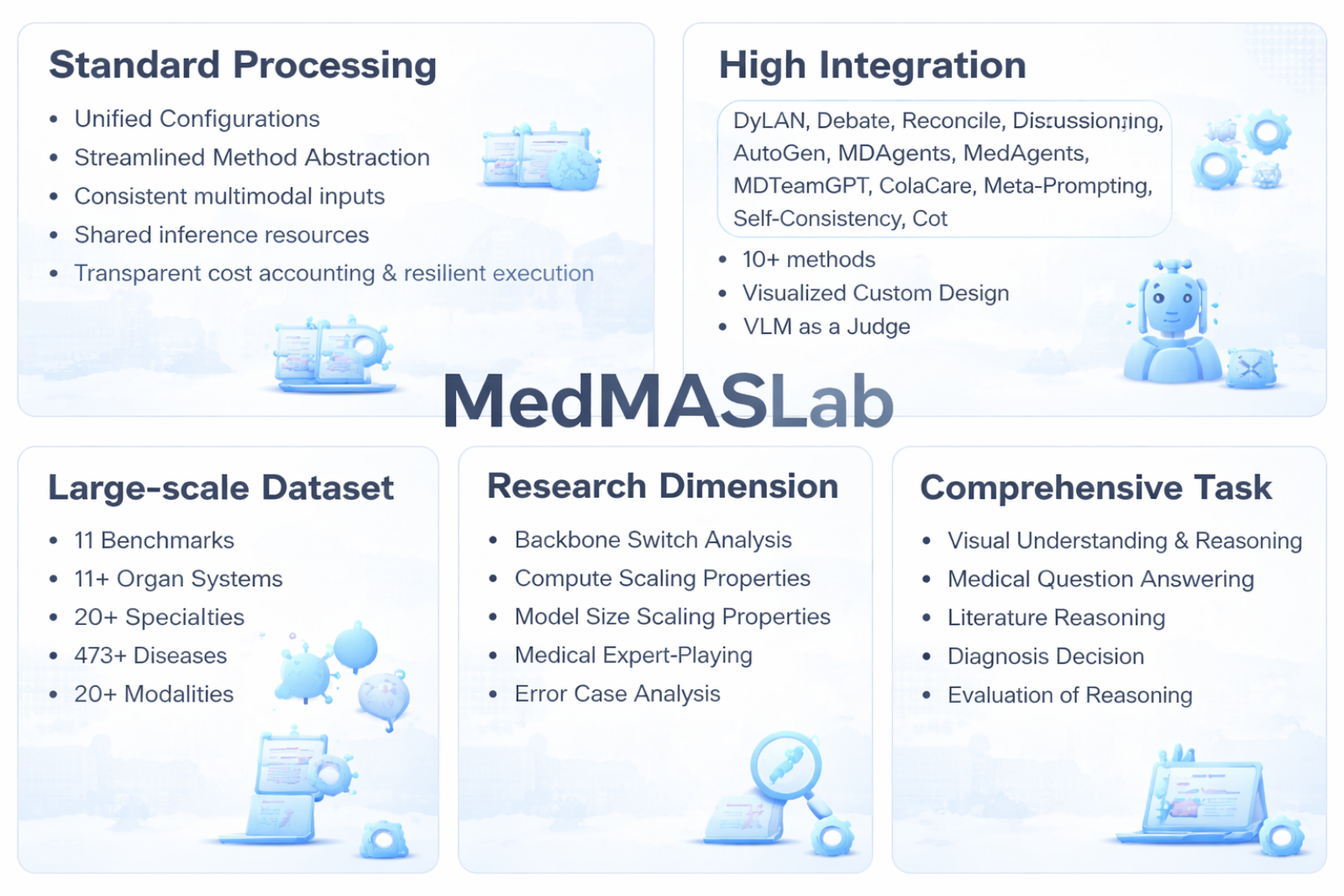}
    \caption{MedMASLab, the first unified orchestration framework designed for medical visual-language multi-agent systems.}
    \label{fig:zongtu}
\end{figure}
\vspace{2em}
\section{Introduction}
\label{sec:intro}
Large Vision-Language Models (LVLMs) have achieved remarkable progress in cross-modal reasoning and zero-shot visual understanding~\cite{bai2025qwen3,liu2024improved}. Despite these gains, the deployment of LVLMs in clinical environments remains constrained by the semantic gap between general-purpose pre-training distributions and the high-fidelity, specialized requirements of medical diagnostics. While prompt engineering has yielded marginal improvements, single-model architectures remain susceptible to stochastic hallucinations and a lack of verifiable logic when processing high-dimensional medical modalities~\cite{huang2025survey,min2023factscore}. The emerging paradigm of Multi-Agent Systems (MAS) offers a collaborative framework to mitigate these failures by decomposing complex diagnostic tasks into specialized, communicative sub-modules~\cite{hu2025landscape,comanici2025gemini}. However, the current medical MAS landscape is characterized by extreme \emph{architectural fragmentation}. Existing systems ~\cite{chen2025mdteamgpt,kim2024mdagents,wang2025colacare,suzgun2024meta,tang2024medagents} are often task-specific, utilizing non-standardized communication protocols and idiosyncratic preprocessing pipelines that lack cross-specialty generalizability. This fragmentation leads to three critical bottlenecks: \textbf{(1) Evaluative heterogeneity and metric failure.} The absence of unified benchmarking environments prevents fair comparative analysis across diverse MAS architectures. Traditional rule-based metrics (e.g., exact string matching) fail to capture the nuance of clinical reasoning, often penalizing valid diagnostic variations. \textbf{(2) Structural opacity and error propagation.} In complex collaborative architectures such as those simulating Multidisciplinary Team (MDT) consultations, the lack of standardized auditing tools makes it difficult to trace information cascades and identify the root cause of systemic failure. \textbf{(3) Architectural integration friction.} The exponential complexity involved in scaling MAS to accommodate hundreds of diseases and heterogeneous imaging modalities (e.g., CT, MRI, Pathology) creates significant technical barriers, impeding the transition from experimental prototypes to clinical-grade systems.

To bridge these gaps, we propose \textbf{MedMASLab}, the first unified, comprehensive, and user-friendly platform specifically engineered for medical visual-language multi-agent systems. In summary, our contributions are:
\begin{itemize}[label=\textbullet,nosep,leftmargin=*]
   \item We introduce a highly composable orchestration framework that abstracts inter-agent communication from modality-specific feature extraction. MedMASLab successfully integrates \textbf{11} heterogeneous MAS methods across \textbf{24} medical modalities and \textbf{473} diseases, establishing a standardized computational manifold for clinical agents. 
   \item We systematically expose the severe vulnerabilities of traditional rule-based metrics in medical MAS assessment, demonstrating that rigid string matching penalizes formatting variations rather than genuine reasoning flaws. To resolve this, we establish a \textbf{zero-shot, multimodal-aware semantic evaluation paradigm} that ensures fair, format-agnostic diagnostic verification.
   \item Through a comprehensive comparative analysis of integrated MAS methods, we provide a rigorous quantification of the \textbf``{specialization Penalty''}. Our findings reveal that current architectures suffer from high task-specificity; we characterize the Pareto frontier between agent communication complexity, inference cost, and clinical robustness.
\end{itemize}
\vspace{1.5em}
\section{Related Work}
\newcommand{\cnum}[1]{\textcircled{\scriptsize #1}}
\begin{table*}[h]
\centering
\normalsize
\setlength{\tabcolsep}{2.5pt}
\renewcommand{\arraystretch}{1.05}

\caption{\textbf{Taxonomy of MAS methods and their design dimensions:} \textbf{1.\,Interaction:} Inter-agent communication pattern. \textbf{2.\,Role:} Fixed vs.\ dynamic agent identities. \textbf{3.\,Tool:} Yes/No. \textbf{4.\,Adaptivity:} Feedback-driven self-tuning. \textbf{5.\,Decision:} Consensus rule. \textbf{6.\,Retrieval:} Access to external knowledge/search.}
\label{tab:agent_taxonomy}
\resizebox{\textwidth}{!}{%
\begin{tabular}{@{}c l l c c c l l@{}}
\toprule
\textbf{No.} & \textbf{Method} & \textbf{Interaction} & \textbf{Role} &
\textbf{Tool} & \textbf{Adaptivity} & \textbf{Decision} &
\textbf{Retrieval} \\
\midrule

\multicolumn{8}{@{}l}{\textbf{Single-Agent Baselines}}\\
\cnum{1}  & VLM (Vanilla)    & Single        & Fixed   & No  & No               & Direct Output             & No \\
\cnum{2}  & CoT~\cite{wei2022chain}              & Single        & Fixed   & No  & No               & Direct Output             & No \\
\midrule

\multicolumn{8}{@{}l}{\textbf{Multi-Agent Systems for General Tasks}}\\
\cnum{3}  & Self-Consistency~\cite{wang2022self} & Independent   & Fixed   & No  & No               & Voting                    & No \\
\cnum{4}  & Debate~\cite{du2024improving}           & Debate              & Fixed   & No  & Iterative              & Consensus                 & No \\
\cnum{5}  & Discussion~\cite{lu2024llm}       & Discussion          & Fixed   & No  & Iterative              & Consensus                 & No \\
\cnum{6}  & Reconcile~\cite{chen2024reconcile}        & Round-Table         & Fixed   & No  & Iterative              & Consensus                 & No \\
\cnum{7}  & DyLAN~\cite{liu2024dynamic}            & Dynamic Graph       & Dynamic & No  & Iterative              & Aggregation               & No \\
\cnum{8}  & AutoGen~\cite{wu2024autogen}          & Conversational      & Dynamic & Yes & Iterative              & Termination Condition     & Feedback Driven \\
\cnum{9}  & Meta-Prompting~\cite{suzgun2024meta}   & Hub-and-Spoke       & Dynamic & Yes & Iterative              & Synthesis                 & No \\
\midrule

\multicolumn{8}{@{}l}{\textbf{Medical-Specific Multi-Agent Systems}}\\
\cnum{10} & MedAgents~\cite{tang2024medagents}        & Collaborative         & Dynamic & No  & Iterative              & Consensus                 & No \\
\cnum{11} & MDAgents~\cite{kim2024mdagents}         & Adaptive              & Dynamic & No  & Complexity-Aware       & Consensus                 & No \\
\cnum{12} & MDTeamGPT~\cite{chen2025mdteamgpt}        & Residual Discussion   & Dynamic & Yes & Self-Evolving          & Consensus                 & Experience KB \\
\cnum{13} & ColaCare~\cite{wang2025colacare}         & Blackboard            & Fixed & No  & Conflict Resolution    & Consensus                 & Medical KB \\
\cnum{14} & LINS~\cite{wang2025lins}             & Iterative Pipeline    & Fixed   & No  & Iterative Retrieval    & Citation Synthesis        & PubMed DB \\
\cnum{15} & MedAgentAudit~\cite{gu2025medagentaudit}    & Monitored Discussion  & Fixed   & No  & Failure Diagnosis      & Error Quantification      & No \\
\cnum{16} & MedLA~\cite{ma2025medla}            & Logic-Graph Based     & Dynamic & No  & Self-Correction        & Logical Consensus         & Medical KB \\
\cnum{17} & CXRAgent~\cite{lou2025cxragent}         & Director-Orchestrated & Dynamic & Yes & Multi-Stage Refinement & Orchestrated Finalization & No \\
\cnum{18} & MoMA~\cite{gao2025moma}             & MoE-Inspired          & Dynamic & No  & Feature-Adaptive       & Gated Aggregation         & No \\
\cnum{19} & MedOrch~\cite{he2025medorch}          & Mediator-Guided       & Dynamic & No  & Mediator-Refinement    & Mediator-Synthesis        & No \\
\bottomrule
\end{tabular}%
}
\vspace{0.5em}

\end{table*}

\vspace{-0.5em}
\noindent \textbf{General MAS.}
To address highly complex tasks, the research paradigm has shifted towards multi-agent systems. Previous studies utilize multi-path sampling~\cite{wang2022self}, enhance consensus and error correction mechanisms~\cite{du2024improving,lu2024llm,chen2024reconcile}, and achieve competitive performance in dynamic task allocation and topological structure~\cite{wu2024autogen,liu2024dynamic}. Although these general frameworks significantly improve the model's ability to complete tasks, they generally face the challenge of balancing universality and domain specificity, lacking in-depth integration of professional domain knowledge. Especially in medical scenarios where high-risk multimodal decision-making is required, their effectiveness and safety still face challenges.

\noindent \textbf{Medical MAS.}
Inspired by the advancements in general multi-agent systems, research in the medical domain has actively adopted agent collaboration paradigms. These efforts primarily focus on simulating Multidisciplinary Team (MDT) consultations~\cite{tang2024medagents,chen2025mdteamgpt}, constructing complex medical workflow orchestration~\cite{suzgun2024meta,chen2025mediator,kim2024mdagents,ma2025medla,wang2025lins}, developing specialized auditing and verification mechanisms~\cite{gu2025medagentaudit,wang2025colacare}, and integrating visual information~\cite{gao2025moma,lou2025cxragent}. Despite notable progress, these medical-specific MAS suffer from critical limitations. Their capabilities often remain confined to specific modalities and tasks, lacking generalizability to broader multimodal clinical scenarios. Furthermore, the absence of unified frameworks leads to fragmented codebases and data processing pipelines, making fair evaluation and method reproduction exceedingly difficult.

\noindent \textbf{Unified Medical Orchestration.} To rectify the lack of standardized infrastructure, MedMASLab introduces a unified orchestration framework that abstracts the core primitives of medical MAS. Unlike previous task-specific codebases, our platform decouples the agent logic from the underlying model, enabling the first large-scale evaluation of heterogeneous methods in a common multimodal manifold. By substituting traditional text backbones with Large Vision-Language Models (LVLMs), we extend the utility of classic MAS topologies to the multimodal clinical domain. As categorized in Table~\ref{tab:agent_taxonomy}, MedMASLab provides a standardized environment for evaluating interaction topologies, role dynamics, and decision-making logic across a spectrum of clinical benchmarks.
\vspace{1.5em}
\input{MAS/inference_and_evaluation}
\input{study}
\section{Conclusion}
MedMASLab, as a unified platform for medical vision-language multi-agent systems, addresses critical challenges currently faced in the medical domain. To establish a rigorous testbed, the inference subsystem standardizes method abstractions and multimodal inputs, thereby eliminating confounding engineering variables. (1) MedMASLab comprehensively tackles the challenges of medical data. It supports 473 diseases and 24 data modalities, effectively addressing the high-dimensional heterogeneity of medical data. By supporting local deployment, MedMASLab ensures the security of privacy-sensitive data. Its visualized evaluation mechanism guarantees the interpretability of decision-making processes. (2) MedMASLab utilizes VLM-based semantic judgment to assess diagnostic correctness. Our systematic analysis of five evaluation protocols demonstrates that rigid rule-based matching penalizes formatting variations rather than genuine clinical reasoning flaws. This approach overcomes severe shortcomings present in traditional rule-based metrics and can provide fairer evaluation results for medical multi-agent systems. (3) Extensive evaluation and ablation studies reveal that current medical multi-agent systems exhibit high task-specificity but limited generalizability. Achieving optimal performance in the medical domain requires customized agent design, optimized interaction mechanisms, and deep integration with domain-specific knowledge. Furthermore, the inherent capabilities of backbone models significantly influence system stability and resource consumption. The platform also includes an interactive graphical user interface to lower development barriers and facilitate rapid experimentation and system comparison.
\bibliography{references}
\bibliographystyle{plain}
\appendix
\section*{Appendix}
\input{Experiment_Details}
\begin{figure}[h]
\vspace{-2em}
    \centering
    \includegraphics[width=\linewidth, height=0.3\textheight]{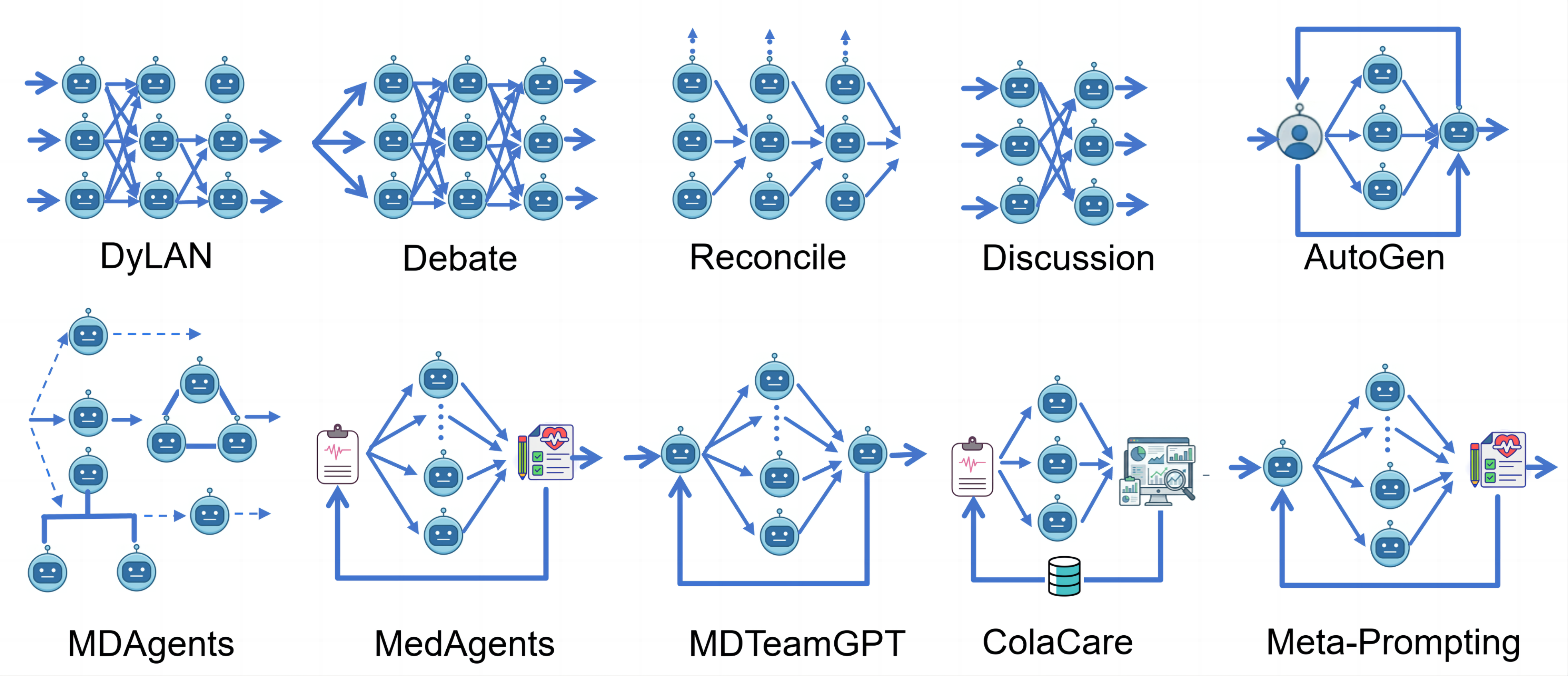}
    \caption{Simplified framework diagrams of DyLAN, Debate, Reconcile, Discussion, AutoGen, MDAgents, MedAgents, MDTeamGPT, ColaCare, and Meta-Prompting.}
    \label{Figure:10MAS}
\end{figure}
\input{MAS/methods}
\input{error}
\input{MAS/gui}

\end{document}

%% file: MAS/inference_and_evaluation.tex
\section{MedMASLab: A Unified Orchestration Framework}

To mitigate architectural fragmentation and evaluative inconsistency in medical Multi-Agent System (MAS) research~\cite{hu2025landscape}, we propose \textbf{MedMASLab}. The framework is engineered as a decoupled orchestration layer that bridges the gap between raw multimodal clinical signals and high-order reasoning agents. MedMASLab is structured into two core subsystems: (1) a standardized inference environment for unified agent execution and (2) a multimodal semantic verification engine designed to quantify diagnostic fidelity beyond surface-level overlap.

\subsection{Multimodal Agentic Orchestration}
\label{sec:inference}

Existing medical MAS often rely on idiosyncratic I/O logic and heterogeneous preprocessing pipelines, which introduce confounding variables during benchmarking. We formalize the orchestration process through a unified abstraction of agentic workflows. MedMASLab eliminates implementation bias via five core design principles: \textit{streamlined method abstraction}, \textit{consistent multimodal inputs}, \textit{shared inference resources}, \textit{unified configurations}, and \textit{transparent cost accounting}.

\begin{wrapfigure}{r}{0.5\textwidth}  
  \vspace{-1.5em}
  \centering
  \includegraphics[width=\linewidth]{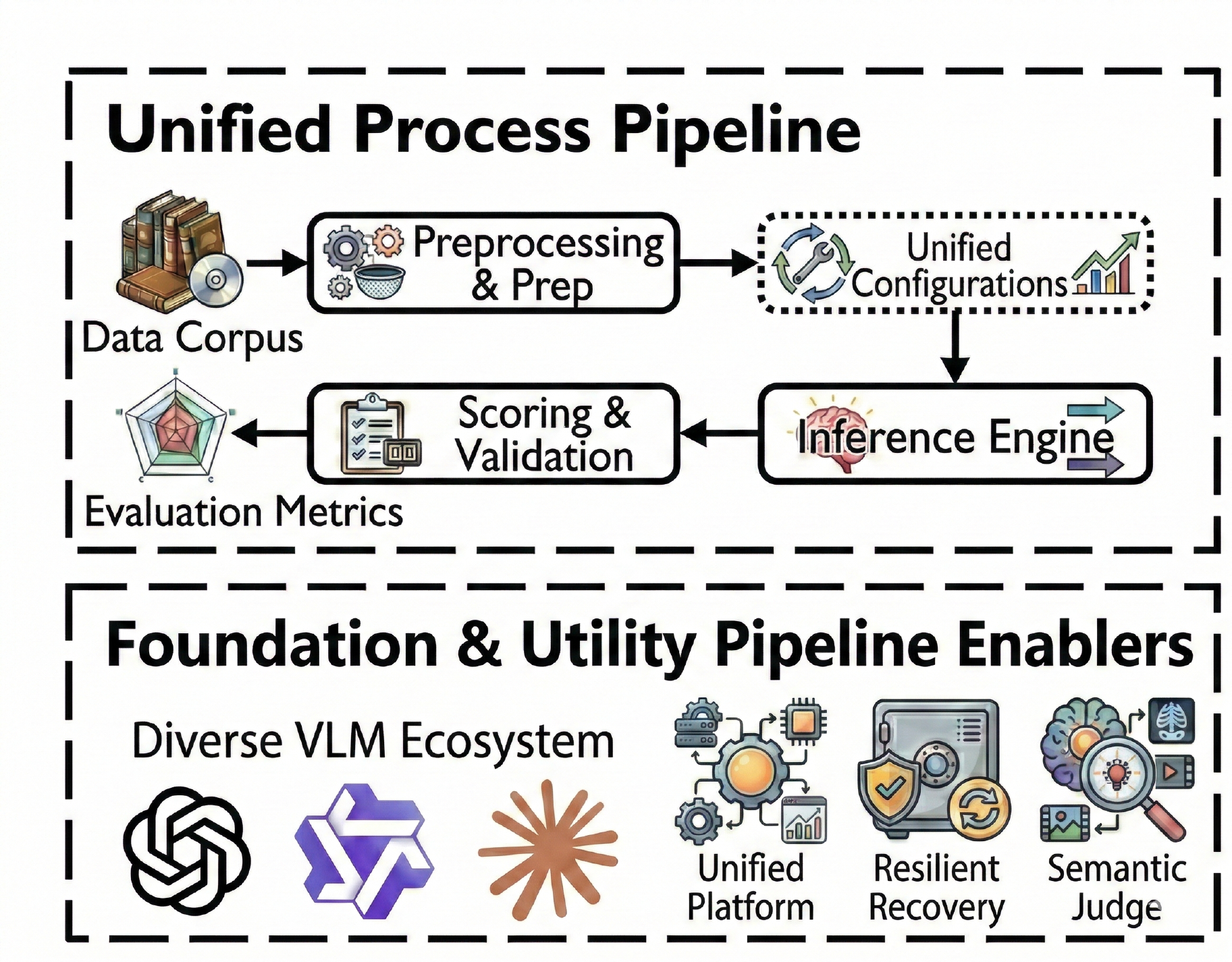}
  \captionof{figure}{Framework of MedMASLab.}
  \label{Figure:framework}
  \vspace{-2em}
\end{wrapfigure}

\noindent
\textbf{Streamlined method abstraction.}
MedMASLab unifies diverse collaboration paradigms (\textit{e.g.,} Chain-of-Thought~\cite{wei2022chain}, debates~\cite{liu2024dynamic,du2024improving}, hierarchical coordination~\cite{suzgun2024meta,kim2024mdagents}, MDT simulations~\cite{wang2025colacare}). Methods must expose a single inference function returning a standard tuple: $\mathcal{R} = (\mathbf{y}, \Gamma, \Theta$), where $\mathbf{y}$ is the medical response, $\Gamma$ represents token usage metrics, and $\Theta$ denotes the specific agentic topology configuration. This minimal Pythonic interface preserves internal agent logic and multi-turn topologies while orchestrating outputs through a single, reproducible code path.

\noindent
\textbf{Consistent multimodal inputs.}
Medical benchmarks span pure-text (Med-QA~\cite{jin2021disease}), images (SLAKE~\cite{liu2021slake}), and videos (MedVidQA~\cite{gupta2023dataset}). To prevent preprocessing from skewing results, a centralized \textit{dataset registry} generates uniform representations containing standardized options, media paths, answer types, and evaluation flags. An adaptive sampler extracts key video frames within a configurable budget to bound visual token consumption. This method-transparent preprocessing ensures observed differences reflect reasoning capabilities, not data intake disparities.

\noindent
\textbf{Shared inference resources.}
Infrastructure heterogeneity invalidates fair comparisons. MedMASLab unifies inference via a shared \textit{dynamic vLLM~\cite{kwon2023efficient} serving layer} that automatically allocates GPUs, provisions instances, and configures context windows to prevent conflicts. Methods communicate with identical model instances via an OpenAI-compatible API, ensuring uniform weights and hardware. This API-level decoupling renders VLM selection orthogonal to MAS logic, allowing seamless backbone model swapping (\textit{e.g.,} LLaVA~\cite{liu2024improved}, Qwen~\cite{bai2025qwen3}) without code modifications to validly assess algorithmic superiority.

\noindent
\textbf{Unified configurations.}
MedMASLab strictly isolates \textit{method-specific algorithmic parameters} (\textit{e.g.,} DyLAN debate rounds) from \textit{infrastructure parameters} (\textit{e.g.,} context window, batch size), preventing unfair advantages. Infrastructure parameters apply uniformly via the serving layer, while algorithmic parameters are externalized in configuration files, facilitating robust ablation studies without source-code changes and directly linking agent topologies to clinical performance.

\noindent
\textbf{Transparent cost accounting \& resilient execution.}
Balancing accuracy with compute costs is vital in clinical settings. MedMASLab natively tracks every VLM call, aggregating tokens and latency into a per-sample structured ledger. Summary statistics enable fine-grained cost-accuracy analyses, identifying diminishing returns from extra deliberation rounds. To mitigate transient failures during large-scale campaigns, MedMASLab employs continuous JSONL checkpointing, restart-triggered deduplication, and an auto-cleansing module that re-queues corrupted entries, ensuring unbiased, complete results without full restarts.

\vspace{1em}
\subsection{Rethinking MAS Evaluation: From Rules to Semantics}
\label{sec:evaluation}

Evaluation misclassifications in medical AI risk patient safety and obscure methodological progress. Traditional metrics often rely on fragmented string matching, which fails to capture the nuance of clinical reasoning. MedMASLab replaces these brittle techniques with a unified VLM-based evaluation paradigm, where a dedicated high-capacity judge model assesses diagnostic fidelity based on semantic intent rather than superficial lexical overlap.
\vspace{1em}
\subsubsection{From Rule-Based Matching to Semantic Verification.} 
Rule-based evaluation is inherently brittle, often penalizing correct reasoning expressed in non-canonical terminology. To quantify this bias, we implement a spectrum of five evaluation protocols, ranging from high-level semantic reasoning to rigid character-level identity: \textbf{VLM-SJ}, namely Semantic Judge, employs a zero-shot Qwen2.5-VL-32B-Instruct~\cite{bai2025qwen3} to directly assess semantic equivalence between verbose outputs and the ground truth, requiring no task-specific fine-tuning; \textbf{VLM-EC}, or Extract-Compare, uses an VLM solely to isolate a canonical option letter from free-form text before applying string matching. The remaining three protocols are rule-based: \textbf{Rule-MR}, representing Multi-Regex, applies a prioritized cascade of regular expressions—such as bracketed options and specific answer prefixes—to capture common formatting patterns; \textbf{Rule-FL}, or First-Letter, naively extracts the initial valid A–E character; and \textbf{Rule-EM}, denoting Exact Match, demands absolute character-level identity.

\begin{figure}[t]
    \centering
    \includegraphics[width=1\textwidth]{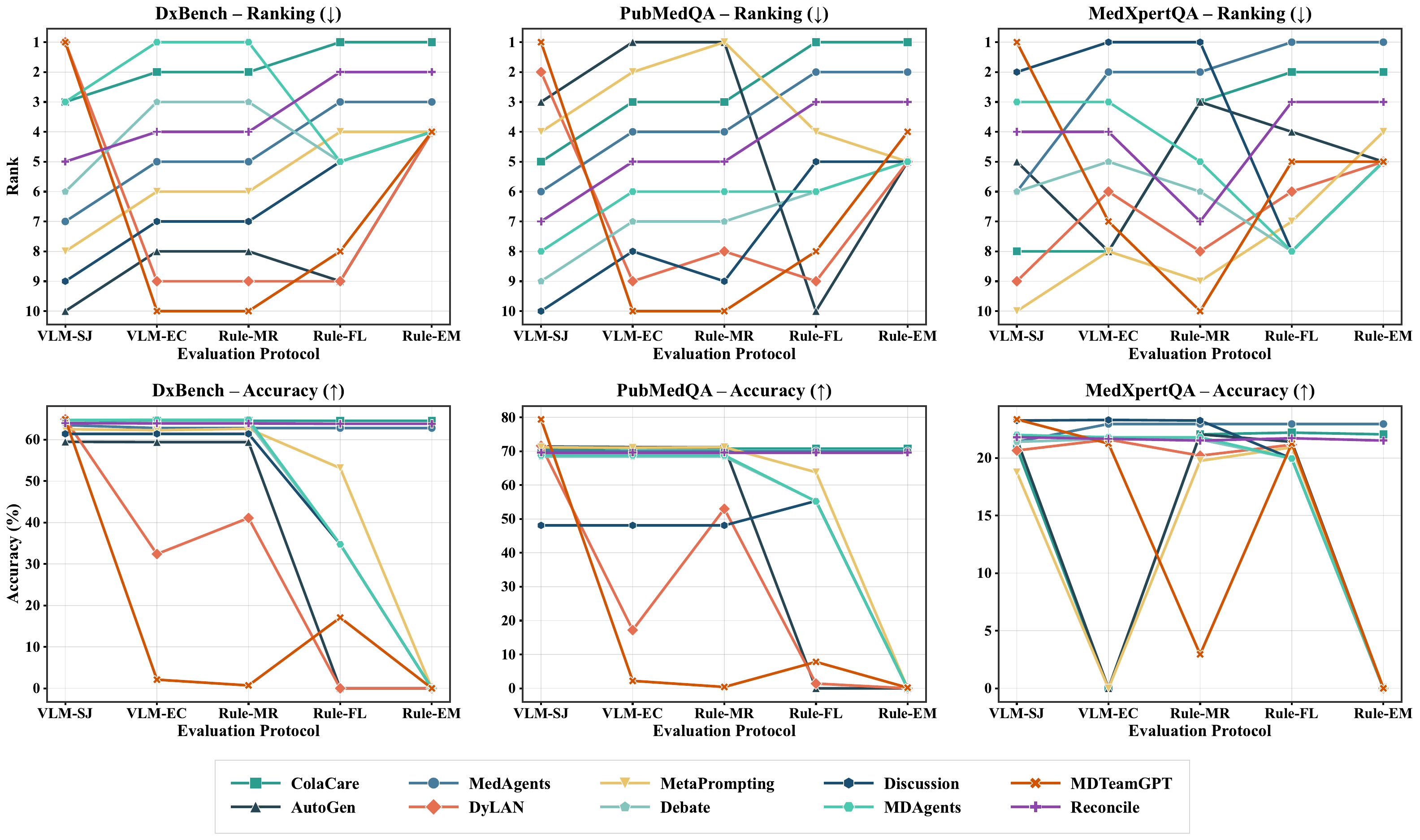}
    \vspace{-4pt}
    \caption{Performance and ranking variations of different MAS methods across five evaluation protocols on DxBench, PubMedQA, and MedXpertQA.}
    \label{fig:eval_protocols}
\end{figure}

Empirical results across three benchmarks (Fig.~\ref{fig:eval_protocols}) demonstrate that protocol selection drastically alters both absolute scores and relative rankings. On PubMedQA, MDTeamGPT~\cite{chen2025mdteamgpt} ranks 1st under VLM-SJ (79.40\%) but collapses to last place (0.40\%) under Rule-MR; DyLAN~\cite{liu2024dynamic} plummets from 71.60\% to 0\% under Rule-EM. 
Crucially, MedXpertQA reveals \textbf{non-monotonic performance fluctuations} with protocol-specific sensitivities rather than linear declines. AutoGen~\cite{wu2024autogen}, ColaCare~\cite{wang2025colacare}, and MetaPrompting~\cite{suzgun2024meta} drop to near 0\% under VLM-EC but recover to >20\% under Rule-MR, as verbose reasoning confounds extractors while regex coincidentally captures options. Conversely, MDTeamGPT survives VLM-EC (22.20\%) but fails Rule-MR (2.90\%), as its output defies regex but remains VLM-comprehensible. Nearly all methods crash to 0\% under Rule-EM, proving complex MAS workflows resist isolated option generation.

This systemic degradation stems from instruction-following fatigue. Frameworks utilizing lengthy deliberation chains often lose formatting adherence over multiple rounds, despite reaching accurate clinical conclusions. Traditional metrics thus unfairly penalize formatting violations over genuine reasoning flaws. Conversely, methods enforcing concise outputs (ColaCare~\cite{wang2025colacare}, Reconcile~\cite{chen2024reconcile}) maintain stable accuracies. In verbose MAS workflows, rigid evaluation protocols can introduce format-induced noise and artificially skew performance rankings. To neutralize this format-induced noise, MedMASLab adopts VLM-SJ as the default for all formal evaluations, ensuring a rigorous, format-agnostic assessment of methodological efficacy.

\noindent
\textbf{Multimodal-aware evaluation.}
In the medical domain, a textually coherent response may remain perceptually ungrounded, potentially contradicting pathological features present in the source scans. To resolve this, MedMASLab provides the judge model with the \textbf{identical multimodal context} including high-resolution radiographs and temporal video frames, provided to the agents during inference. This visual grounding ensures that the evaluation is not merely a linguistic check, but a verification that the agent's reasoning is consistent with the primary visual evidence.

\noindent
\textbf{Holistic performance profiling.}
Since single metrics often obscure practical utility, MedMASLab logs a structured JSON \textbf{holistic performance profile} (correctness, latency, VLM calls, tokens, configurations) per sample. Cross-referencing metadata enables nuanced analytics, precisely decoupling \textbf{system-level failures} (\textit{e.g.}, protocol breakdowns) from \textbf{semantic clinical errors} (\textit{e.g.}, flawed reasoning), preventing misattribution and optimizing deployment.

\noindent
\textbf{Robustness and evaluation transparency.}
Meticulous prompt engineering directs judge attention to semantic content rather than surface formatting, comparing outputs against both labels and content. Ambiguous verdicts or API errors are explicitly flagged. Alongside auto-cleansing and persistent records, the pipeline guarantees full auditability and arbitrary-granularity replication.

%% file: study.tex
\begin{table}[t!]
  \caption{We compare the performance of general-task and medicine-specific 
  methods across five aspects in the medical domain (\hlML{Medical Literature 
  Reasoning}, \hlVU{Medical Visual Understanding and Reasoning}, \hlQA{Medical 
  Question Answering}, \hlDD{Diagnosis Decision}, and \hlMM{Evaluation of 
  Medical Reasoning Chains}). Avg-V denotes the average accuracy 
  ($\uparrow$). \textbf{Best} and \underline{second-best} numbers are highlighted.}
  \vspace{3mm}
  \label{tab:general-methods}
  \centering
  \resizebox{\textwidth}{!}{
  \begin{tabular}{l|
    >{\columncolor{colorML}}c|        
    >{\columncolor{colorQA}}c
    >{\columncolor{colorQA}}c
    >{\columncolor{colorQA}}c
    >{\columncolor{colorQA}}c|        
    >{\columncolor{colorVU}}c
    >{\columncolor{colorVU}}c
    >{\columncolor{colorVU}}c
    >{\columncolor{colorVU}}c|        
    >{\columncolor{colorDD}}c|        
    >{\columncolor{colorMM}}c|        
     >{\columncolor{colorAvg}}c}
    \toprule
    Method & PubMedQA & MedQA & MedBullets & MMLU & VQA-RAD & 
    SLAKE-En & MedVidQA & MedCMR & MedXpertQA-MM & DxBench & 
    M3CoTBench & Avg-V \\
    \midrule
    \multicolumn{13}{c}{\textnormal{Qwen-2.5VL-7B-Instruct}} \\
    \midrule
    Single~\cite{bai2025qwen3}        & 68 & 52.8 & 35.7 & 75.2 & 50.4 & 58.3 & 71.6 & 68.1 & 20.8 & 62.9 & 30.8 & 54.1 \\
    Debate~\cite{du2024improving}      & 68.4 & 52.9 & 37.1 & 76.6 & 54.1 & 64.4 & 76.4 & 64.5 & 21.6 & 64.2 & 34 & 55.9 \\
    MDAgents~\cite{kim2024mdagents}    & 68 & 52.3 & 38.4 & 73.9 & 56.6 & 63.8 & \underline{79.1} & \underline{68.9} & 22.6 & 64.7 & \textbf{36.8} & \underline{56.8} \\
    MDTeamGPT~\cite{chen2025mdteamgpt} & \textbf{79.4} & \underline{56.1} & \underline{39} & \textbf{77.6} & 50.3 & 58.3 & 71.6 & 62.7 & \textbf{23.4} & \underline{64.9} & \underline{34.6} & 56.2 \\
    Discussion~\cite{lu2024llm}    & 56 & 52.3 & 35.2 & 74 & \underline{57.3} & \textbf{65.3} & 75 & 65.9 & 23.3 & 61.5 & 31.8 & 54.3 \\
    Reconcile~\cite{chen2024reconcile} & 70.8 & 52.9 & 35.2 & 76 & 54.1 & 58.8 & 71.9 & 66.2 & 22.1 & 63.8 & 30.6 & 54.8 \\
    Meta-Prompting~\cite{suzgun2024meta} & 70.6 & 52.6 & 38 & 73.4 & 51.7 & 58.2 & 78.7 & 61.6 & 21.1 & 64.2 & 29.9 & 54.6 \\
    AutoGen~\cite{wu2024autogen}      & \underline{73} & 50.7 & 36.7 & 73.3 & 56.6 & 62.1 & 77.1 & 67.3 & \underline{23.3} & 61.7 & 28.4 & 55.5 \\
    DyLAN~\cite{liu2024dynamic}        & 62.4 & 53.1 & 35.1 & 75.2 & 47.7 & 58.4 & 69.6 & 64.6 & 21.6 & 63.3 & 33.9 & 53.2 \\
    MedAgents~\cite{tang2024medagents} & 71 & \textbf{56.7} & \textbf{41.9} & 75.3 & 49.5 & 58.9 & 73 & \textbf{72.9} & 21.5 & \textbf{65.2} & 29.2 & 55.9 \\
    ColaCare~\cite{wang2025colacare}   & 71.4 & 54.9 & 38.4 & \underline{77.4} & \textbf{59.5} & \underline{65.2} & \textbf{80.5} & 67.9 & 21.6 & 64.5 & 28.8 & \textbf{57.3} \\
    \midrule
    \multicolumn{13}{c}{\textnormal{LLaVA-v1.6-mistral-7b-hf}} \\
    \midrule
    Single~\cite{liu2024improved}        & 56.6 & 39.2 & 31.2 & 59.9 & 50.8 & 50.7 & 56.1 & 53.3 & 21.8 & 57.6 & 31.9 & 46.3 \\
    Debate~\cite{du2024improving}      & 55 & 43.6 & 33.8 & 59 & \underline{52.8} & 53.1 & 57 & 49.8 & 20.2 & 58.1 & 33.5 & 46.9 \\
    MDAgents~\cite{kim2024mdagents}    & 60.6 & 40.6 & 31.5 & 58.8 & \textbf{54.6} & 53.1 & 64.9 & 52.8 & 21.3 & 54.3 & \textbf{34.8} & 47.9 \\
    MDTeamGPT~\cite{chen2025mdteamgpt} & \underline{65.7} & 41.8 & \textbf{35.8} & \underline{62.4} & 53.2 & 50.9 & 58.5 & 48 & 21.4 & 57.3 & 33.1 & \underline{48} \\
    Discussion~\cite{lu2024llm}    & \textbf{72.3} & 39.8 & 30.2 & 61.9 & 49.3 & 52.8 & 51.4 & 48.3 & 22.1 & 56.5 & 32.3 & 47 \\
    Reconcile~\cite{chen2024reconcile} & 61.8 & \underline{44.5} & 32.6 & 58.3 & 51 & 50.4 & 59.9 & 53.7 & 20.5 & 52.6 & 32.5 & 47.1 \\
    Meta-Prompting~\cite{suzgun2024meta} & 53.4 & 40.8 & 32.2 & 60 & 51.4 & 52.8 & 63.5 & 54.7 & 22.4 & \underline{58.2} & 30.3 & 47.3 \\
    AutoGen~\cite{wu2024autogen}      & 58.1 & 38 & 29.9 & 57 & 51.3 & 50 & \textbf{73.7} & 47 & \textbf{22.7} & 52 & 31 & 46.5 \\
    DyLAN~\cite{liu2024dynamic}        & 44.8 & 37.8 & 30.2 & 58.4 & 50.9 & \textbf{56.4} & 60.9 & \underline{57.2} & 20.4 & 54.2 & 32.5 & 45.8 \\
    MedAgents~\cite{tang2024medagents} & 53.6 & 42.5 & \underline{33.9} & \textbf{63.8} & 48.6 & 51 & 51.4 & 56.1 & 22.2 & \textbf{58.8} & 32 & 46.7 \\
    ColaCare~\cite{wang2025colacare}   & 62.4 & \textbf{46.1} & 31.9 & 58.5 & 52.4 & 51.8 & \underline{73} & \textbf{59.6} & \underline{22.5} & 56.2 & \underline{34.7} & \textbf{49.9} \\
    \bottomrule
  \end{tabular}
  }
\end{table}

\begin{figure}[!h]
    \centering
    \includegraphics[width=\linewidth]{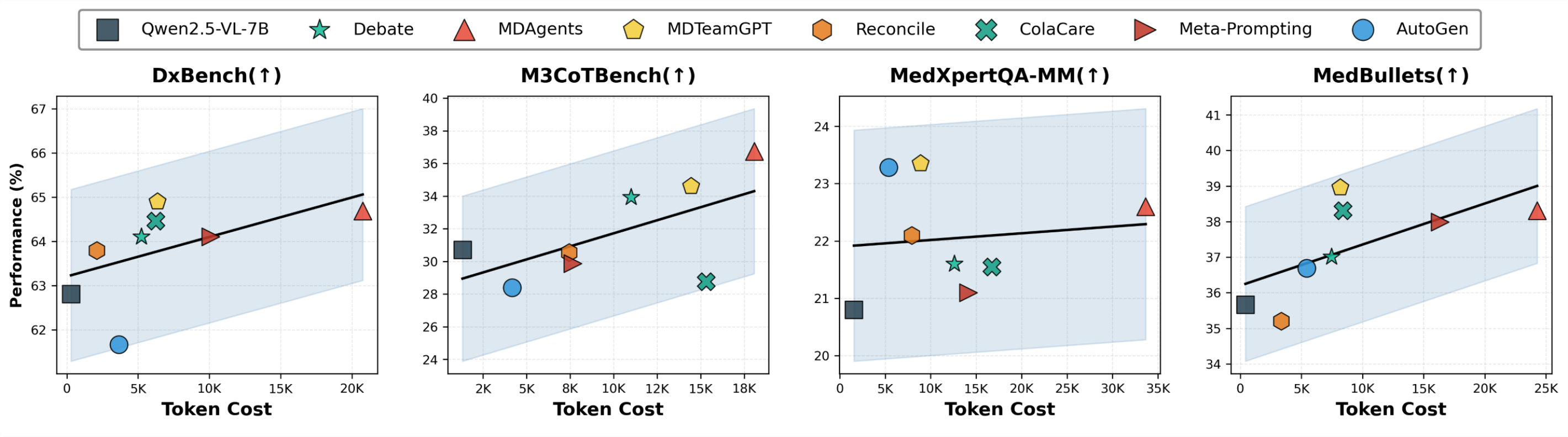}
    \caption{Trade-off between performance and token cost of Qwen2.5VL-7B-Instruct based Multi-Agent Methods across Medical Benchmarks.}
    \label{Figure:performance_vs_tokencost_Qwen7b}
\end{figure}

\vspace{1.5em}
\section{Experiments and Analysis}
In this section, we provide a systematic empirical evaluation of MedMASLab across 11 diverse medical benchmarks. Beyond reporting aggregate performance, we leverage our framework’s decoupled architecture to perform an in-depth analysis of the factors governing agentic efficacy. Specifically, we investigate the interplay between base model configurations and multi-agent collaborative mechanisms across four critical dimensions: (a) Backbone Switch Analysis, (b) Compute Scaling Properties, (c) Model Size Scaling Properties, and (d) Medical Expert-Playing Ablation. 
Finally, we perform a failure mode analysis, utilizing our semantic verification engine to distinguish between instruction-following breakdowns and genuine clinical reasoning errors.

\begin{figure}[!h]
    \centering
    \includegraphics[width=\linewidth]{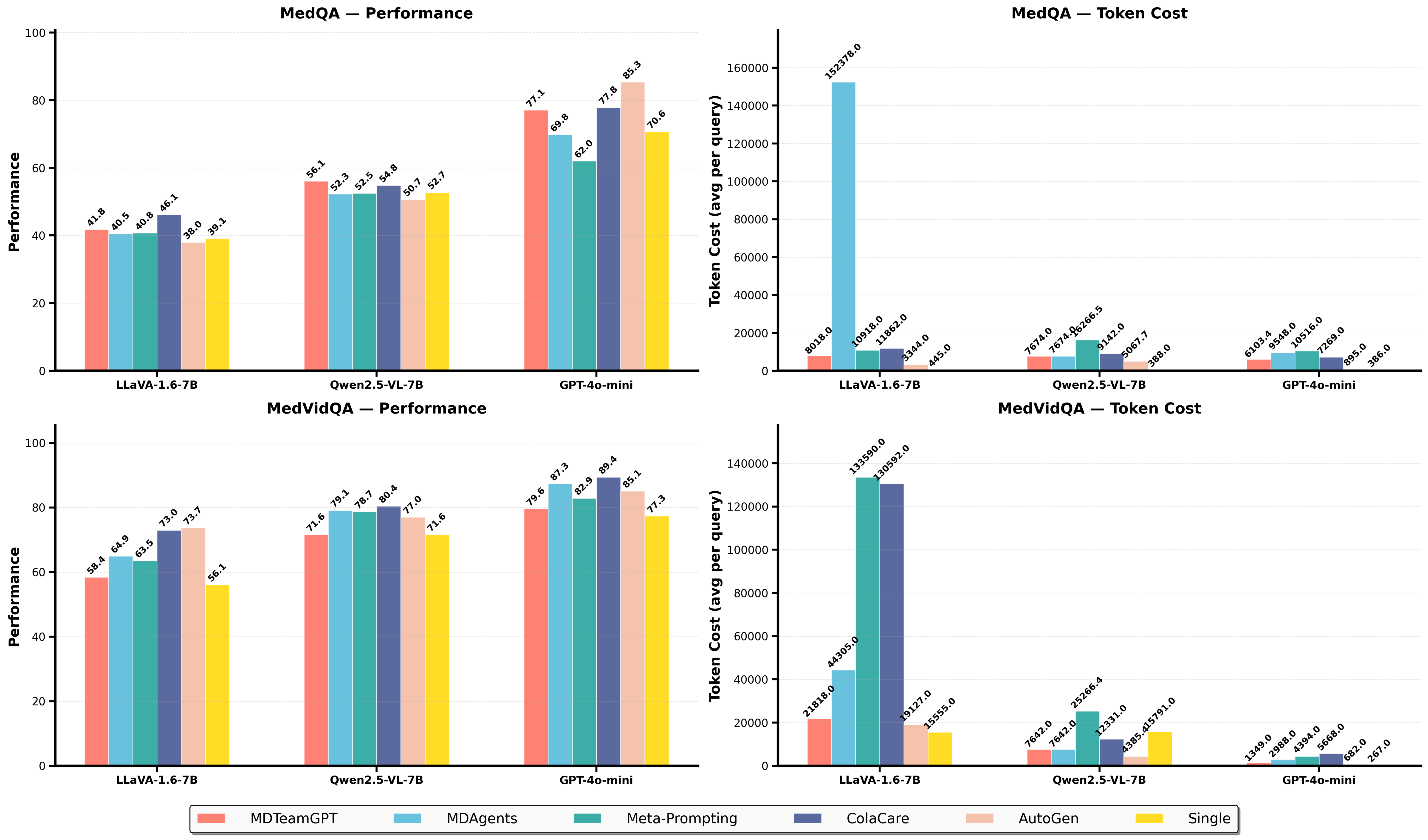}
    \caption{Comparison of Method Performance and Token Cost using Qwen2.5VL-7B, LLaVA-7B, and GPT-4o-mini as Backbones on MedQA and MedVidQA.}
    \label{Figure:qwen_vs_llava}
\end{figure}

\subsection{Experiment Setup}
\label{subsec:Experiment Setup}
\noindent\textbf{Datasets.}
{\sloppy
Based on the specific model capabilities each dataset evaluates, we further group these datasets into five categories: \textbf{Medical Visual Understanding and Reasoning} (Med-CMR~\cite{gong2025med}, SLAKE-En~\cite{liu2021slake}, MedVidQA~\cite{gupta2023dataset}, MedXpertQA-MM~\cite{zuo2025medxpertqa}), \textbf{Diagnostic Decision-Making}(DxBench~\cite{chen2025cod}), \textbf{Medical Literature Reasoning} (PubMedQA~\cite{jin2019pubmedqa}), \textbf{Medical Question Answering}(MedQA~\cite{jin2021disease}, MedBullets~\cite{chen2025benchmarking}, MMLU~\cite{hendrycks2020measuring}, VQA-RAD~\cite{lau2018dataset}), and \textbf{Evaluation of Medical Reasoning Chains} (M3CoTBench~\cite{jiang2026m3cotbench}). Detailed descriptions of each dataset are provided in Appendix A and Table 4.

\noindent\textbf{Implementation Detail.}
Throughout the experimental phase, we primarily utilize the Qwen2.5VL series~\cite{bai2025qwen3}, LLaVA-v1.6~\cite{liu2024improved}, and the GPT series~\cite{achiam2023gpt} as base models. We set the default maximum token limit to 1024 with a temperature of 0.1. To ensure a fair comparison, we restrict these Multi-Agent System frameworks from using any external tools, strictly evaluating them in a zero-shot setting. Each method is required to complete the task using only the inherent capabilities of the backbone Visual Language Model and its multi-agent collaboration mechanism. To ensure the accuracy and reliability of our evaluation, we exclusively use the \textbf{VLM-SJ} (zero-shot Qwen2.5-VL-32B-Instruct semantic judge \cite{bai2025qwen3}) mode throughout the entire experiment.

\begin{figure}[!h]
    \centering
    \includegraphics[width=\linewidth]{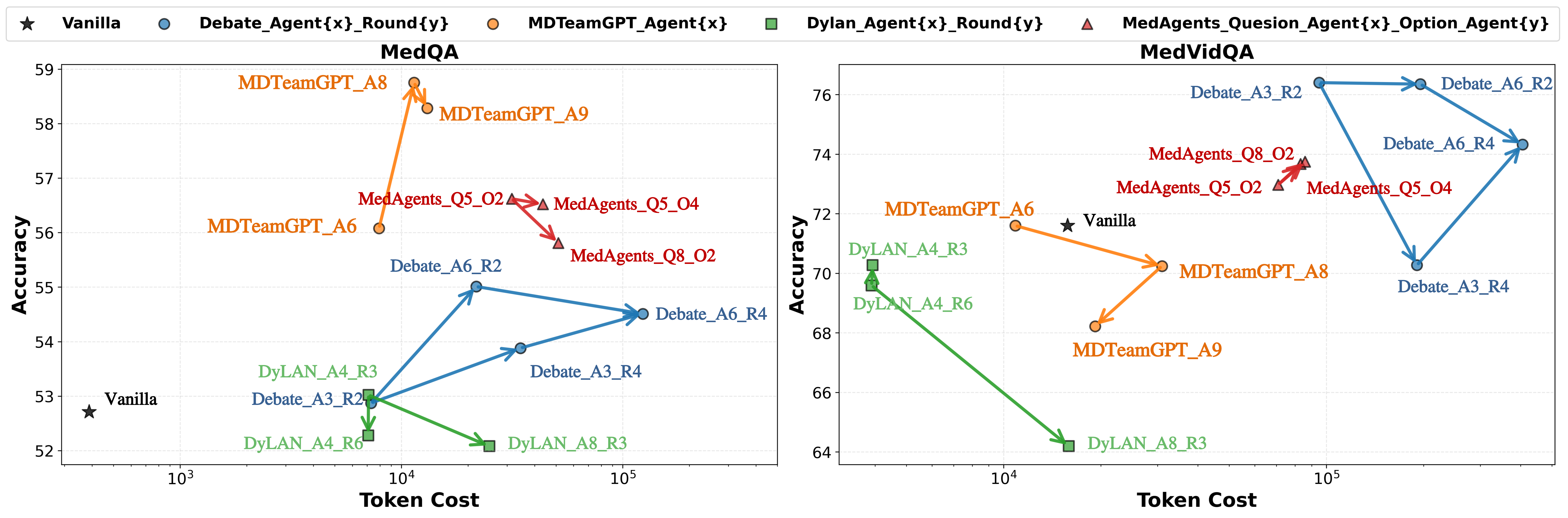}
    \caption{ Trade-off under MAS Compute Scaling on MedQA and MedVidQA.}
    \label{Figure:performance_comparison_colored}
\end{figure}
\vspace{1em}
\subsection{Comparison Experiments}
\noindent\textbf{Performance Across Diverse Medical Benchmarks.} 
We conduct a comprehensive evaluation of both general-purpose and medical-specialized MAS methods across 11 datasets, summarized in Table~\ref{tab:general-methods}. Utilizing both Qwen2.5-VL-7B~\cite{bai2025qwen3} and LLaVA-v1.6-7B~\cite{liu2024improved} as backbones, our results indicate that domain-specific frameworks typically achieve state-of-the-art results on the specific task modalities they are designed for (e.g., diagnostic reasoning vs. visual QA). However, we observe a lack of architectural universality; no single method consistently dominates across all 11 benchmarks. Notably, except for DxBench~\cite{chen2025cod} and M3CoTBench~\cite{jiang2026m3cotbench}, shifting the underlying backbone frequently reorders the method rankings. This suggests that a method’s "collaborative gain" is not an intrinsic algorithmic property but is deeply coupled with the latent reasoning capabilities of the base model.

\noindent\textbf{The Token-Efficiency Frontier.}
To analyze the trade-off between reasoning depth and computational overhead, we plot the performance-vs-token cost distribution in Fig.~\ref{Figure:performance_vs_tokencost_Qwen7b} and Fig.~\ref{fig:performance_vs_tokencost_qwen_7b_merged} (Appendix C). Token consumption serves as a proxy for inter-agent communication density. On datasets requiring high-order cognitive synthesis (e.g., DxBench~\cite{chen2025cod}), we observe a positive correlation where increased token budgets facilitate more rigorous multi-turn deliberation, leading to higher accuracy. Conversely, on retrieval-heavy tasks like MedXpertQA~\cite{zuo2025medxpertqa}, redundant exchanges often introduce semantic noise, resulting in diminishing returns or performance degradation. This highlight that optimal medical MAS design requires task-aware throttling of agent communication to balance diagnostic precision with inference efficiency.

\noindent\textbf{Backbone Switch Analysis.}
We investigate the architectural robustness of MAS frameworks by swapping backbones between Qwen2.5-VL-7B~\cite{bai2025qwen3}, LLaVA-1.6-7B~\cite{liu2024improved}, and GPT-4o-mini~\cite{achiam2023gpt}. As illustrated in Fig.~\ref{Figure:qwen_vs_llava}, while Qwen and GPT-4o-mini maintain stable execution, switching to LLaVA-1.6-7B triggers catastrophic token inflation in specific architectures.
For instance, on MedQA~\cite{jin2021disease}, the token consumption of MDAgents~\cite{kim2024mdagents} escalates to approximately 150,000 tokens per query—a nearly 100$\times$ increase compared to other backbones. This indicates a behavioral instability where weaker instruction-following in the base model prevents agents from reaching ``consensus'' or terminating loops, causing the multi-agent orchestration to degenerate into infinite or redundant dialogue. These findings underscore that the stability of an agentic framework is fundamentally constrained by the instructional grounding of its constituent models.
\par}

\begin{figure}[!h]
    \centering
    \includegraphics[width=0.8\linewidth]{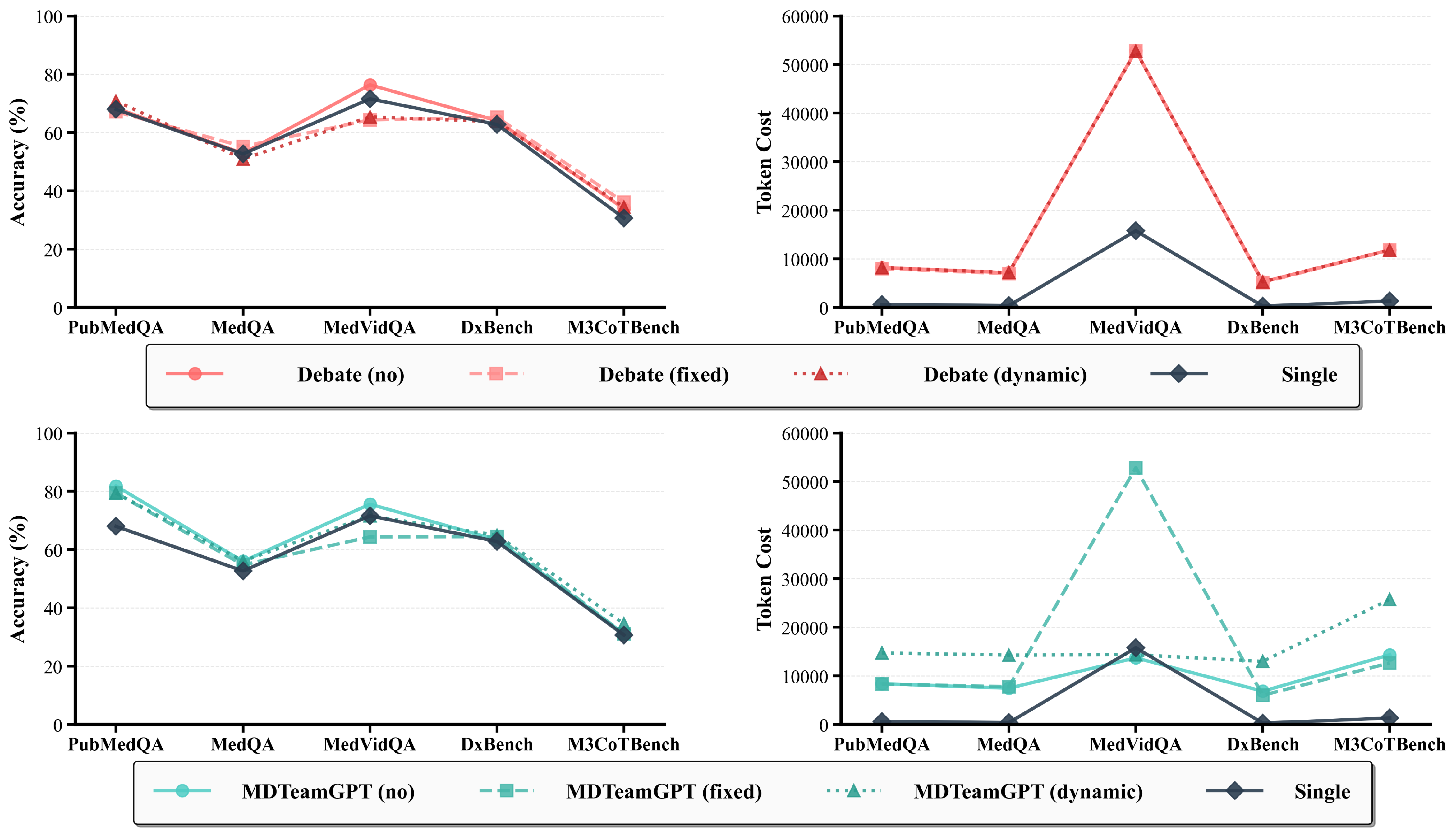}
    \caption{Performance and Token Cost of Debate and MDTeamGPT under Different Medical Expert-Playing Modes.}
    \label{Figure:Debate_MDTeamGPT_merged}
\end{figure}

\noindent\textbf{Compute Scaling Properties.}
We use Qwen2.5VL-7B~\cite{bai2025qwen3} as the base model and compare four configurable methods, including DyLAN~\cite{liu2024dynamic}, Debate~\cite{du2024improving}, MedAgents~\cite{tang2024medagents}, and MDTeamGPT~\cite{chen2025mdteamgpt}, on the MedQA~\cite{jin2021disease} and MedVidQA~\cite{hendrycks2020measuring} benchmarks to investigate the computational scaling properties of Multi-Agent Systems in medical tasks. In Figure~\ref{Figure:performance_comparison_colored}, we observe that simply increasing the number of agents in medical tasks does not always lead to higher accuracy, and there is a balance point between diminishing marginal returns and cost-effectiveness. For instance, on MedQA, MDTeamGPT achieves optimal performance when the number of agents is set to 8, and further increasing the agent scale leads to a decrease in performance. MedQA and MedVidQA exhibit different scaling characteristics, necessitating task-specific parameter configurations and a task-driven scaling strategy. For example, on MedQA, the optimal performance configuration for Debate is Debate-A6-R2, whereas on MedVidQA, the optimal performance configuration for Debate becomes Debate-A3-R2.

\noindent\textbf{Medical Expert-Playing Ablation.}
Medical decision-making is a multifaceted and intricate process. To effectively address these challenges, current medical MAS typically adopt an approach where agents play the role of specialized doctors to simulate real-world expert consultations and multidisciplinary team (MDT) discussions~\cite{kim2024mdagents}. However, this field still lacks a detailed analysis of this medical expert-playing paradigm. To fill this gap, we use Qwen2.5VL-7B~\cite{bai2025qwen3}  as the base model for MDTeamGPT~\cite{chen2025mdteamgpt} and Debate~\cite{du2024improving} to investigate the performance and token cost of MAS frameworks on the PubMedQA~\cite{jin2019pubmedqa}, MedQA~\cite{jin2021disease}, MedVidQA~\cite{hendrycks2020measuring}, DxBench~\cite{chen2025cod}, and M3CoTBench~\cite{jiang2026m3cotbench} datasets under three modes: \textbf{no medical expert-playing} (where the model acts as a General Assistant), \textbf{fixed medical expert-playing}(where the model's roles are fixed), and \textbf{dynamic medical expert-playing}(where the model adaptively assigns its roles based on the input question). We present the comparative results in Figure~\ref{Figure:Debate_MDTeamGPT_merged} and Figure~\ref{Figure:MedAgents_role_playing} (Appendix C). We find that fixed medical expert-playing mode and dynamic medical expert-playing mode lead to a substantial, or even explosive, increase in token cost, but do not necessarily result in an improvement in framework performance. For example, when Debate is configured with fixed medical expert-playing or dynamic medical expert-playing modes, its performance on the M3CotBench dataset slightly increases, yet its performance on the MedVidQA dataset decreases, while the token cost reaches an average of 50,000 tokens per query.

\begin{figure}[!h]
    \centering
    \includegraphics[width=0.8\linewidth]{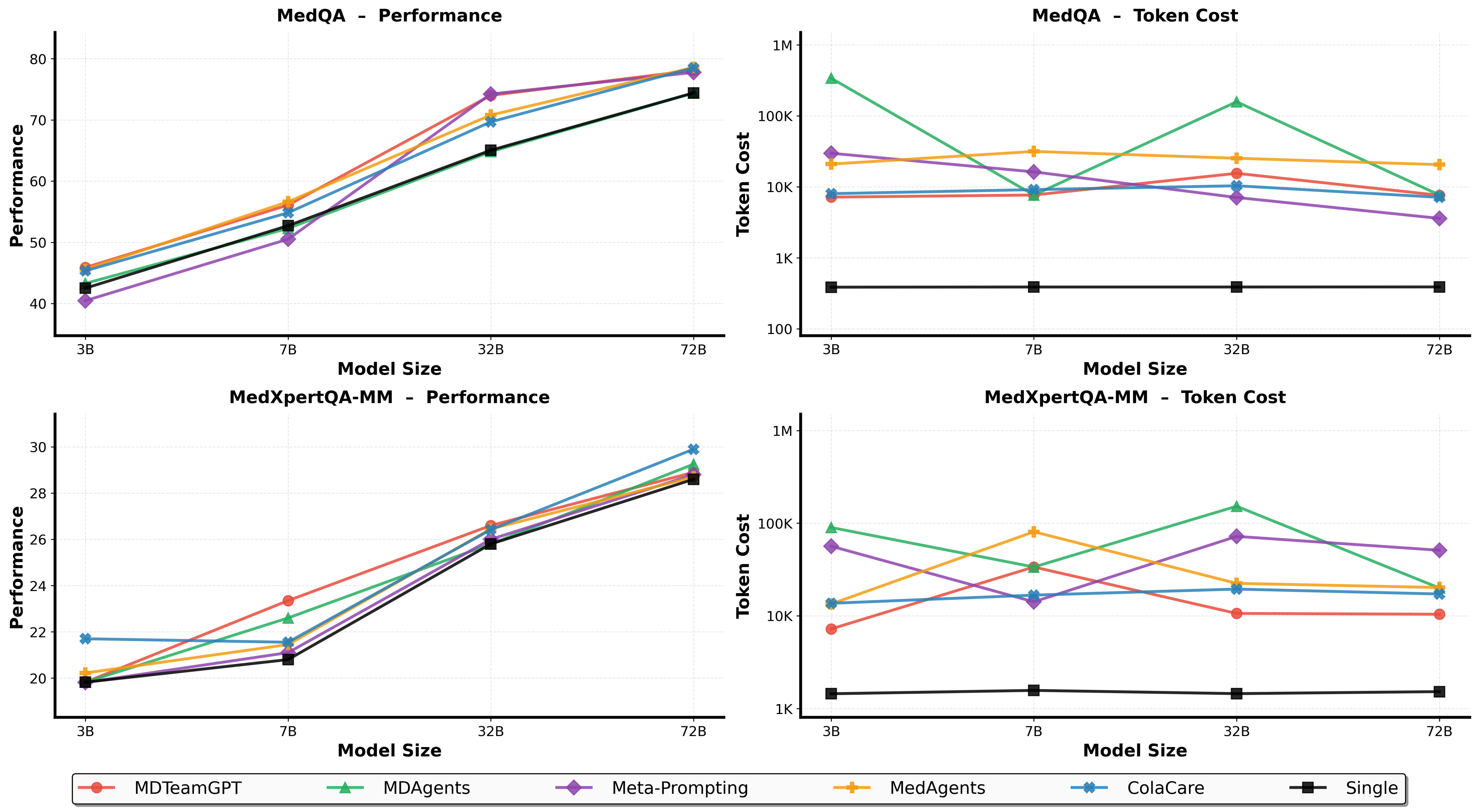}
    \caption{Trade-off under MAS Model Size Scaling on MedQA and MedXpertQA-MM.}
    \label{Figure:model_size_2_model}
\end{figure}

\noindent\textbf{Model Size Scaling Properties.}
We evaluate the performance of four different scales of the Qwen2.5VL series models~\cite{bai2025qwen3} across five representative multi-agent system approaches on MedQA~\cite{jin2021disease} and MedXpertQA-MM~\cite{zuo2025medxpertqa} datasets. This is compared against single-agent baselines to assess the impact of scaling up the backend models. As shown in Figure~\ref{Figure:model_size_2_model}, we observe a significant and continuous upward trend in the accuracy of all multi-agent frameworks and single agents with increasing base model parameter size. On the MedQA dataset, the largest gains from multi-agent collaboration are observed with the 32B model. Conversely, on the MedXpertQA-MM dataset, the greatest benefits from multi-agent collaboration are seen with the 7B model. This suggests that as base models further increase in size, the performance improvement of single agents themselves may accelerate, potentially leading to a flattening or even reduction in the relative marginal gains from multi-agent collaboration. This highlights the importance of considering the cost-effectiveness of multi-agent collaboration based on the characteristics of specific  medical tasks to find the optimal balance between performance and efficiency.

\begin{table}[h]
  \centering
  \normalsize
  \setlength{\tabcolsep}{4pt} 
  \caption{Distribution of outcomes for the Reconcile  framework using Qwen2.5VL series models as the base model. The table compares the proportions of Right Answer, Wrong Answer, Format Error, and Others across multiple medical datasets for 3B and 7B model variants.}
  \label{tab:reconcile-comparison}
  \begin{tabular}{ll|cccc}
    \toprule
    Base Model & Dataset & \begin{tabular}[c]{@{}c@{}}Right\\Answer\end{tabular} & \begin{tabular}[c]{@{}c@{}}Wrong\\Answer\end{tabular} & \begin{tabular}[c]{@{}c@{}}Format\\Error\end{tabular} & Others \\
    \midrule
    \multirow{4}{*}{Qwen2.5VL-3B} & MedVidQA & 40.62\% & 53.38\% & 6.00\% & 0.00\% \\
    & SLAKE-En & 55.54\% & 43.36\% & 0.90\% & 0.20\% \\
    & PubMedQA & 3.00\% & 13.00\% & 84.00\% & 0.00\% \\
    & MedQA & 10.53\% & 14.14\% & 75.33\% & 0.00\% \\
    \midrule
    \multirow{4}{*}{Qwen2.5VL-7B} & MedVidQA & 71.89\% & 28.11\% & 0.00\% & 0.00\% \\
    & SLAKE-En & 58.72\% & 41.28\% & 0.00\% & 0.00\% \\
    & PubMedQA & 70.80\% & 29.20\% & 0.00\% & 0.00\% \\
    & MedQA & 52.88\% & 46.50\% & 0.55\% & 0.07\% \\
    \bottomrule
  \end{tabular}
  \vspace{-1em}
\end{table}

\subsection{Error Analysis}
\label{subsec:Error Analysis}
We further investigate the failure modes of agentic workflows when integrated with VLM backbones. For instance, Reconcile~\cite{chen2024reconcile} utilizes an iterative voting mechanism that necessitates the programmatic extraction of candidate answers from each deliberation round. This architecture imposes a strict requirement for structural consistency; the model must strictly adhere to a predefined format (e.g., JSON) to facilitate reliable parsing. Consequently, any degradation in instruction-following which is frequently observed during complex, multi-turn reasoning, leads to parsing failures that disrupt the voting logic and collapse the final output. From Table~\ref{tab:reconcile-comparison}, it is evident that the base model may fail to generate responses in the required format, leading to errors that impact the final performance assessment of the framework. When utilizing the smaller Qwen2.5VL-3B-Instruct model~\cite{bai2025qwen3}, the Reconcile~\cite{chen2024reconcile} system suffers from formatting errors, particularly on complex, text-heavy reasoning tasks. For example, the format error rate reaches 84.00\% on PubMedQA~\cite{jin2019pubmedqa}, 6.00\% on MedVidQA~\cite{hendrycks2020measuring} and 75.33\% on MedQA~\cite{jin2021disease}. This is because Reconcile’s voting mechanism cannot parse malformed or non-compliant outputs. In contrast, upgrading the base model to Qwen2.5VL-7B fundamentally removes this parsing bottleneck: the format error rates drop to 0.00\% on PubMedQA~\cite{jin2019pubmedqa}, MedVidQA~\cite{hendrycks2020measuring}, and SLAKE-En~\cite{liu2021slake}, and decrease substantially to 0.55\% on MedQA~\cite{jin2021disease}. These results indicate that, in multi-agent systems, failures are primarily driven by breakdowns in communication protocols rather than merely a lack of medical knowledge. Moreover, as the base model scales up and becomes more capable, its outputs become more well-formed, allowing the MAS framework to demonstrate its advantages more clearly.

\begin{wrapfigure}[13]{r}{0.42\textwidth}
  \vspace{-1.3em}
  \centering
  \includegraphics[width=\linewidth]{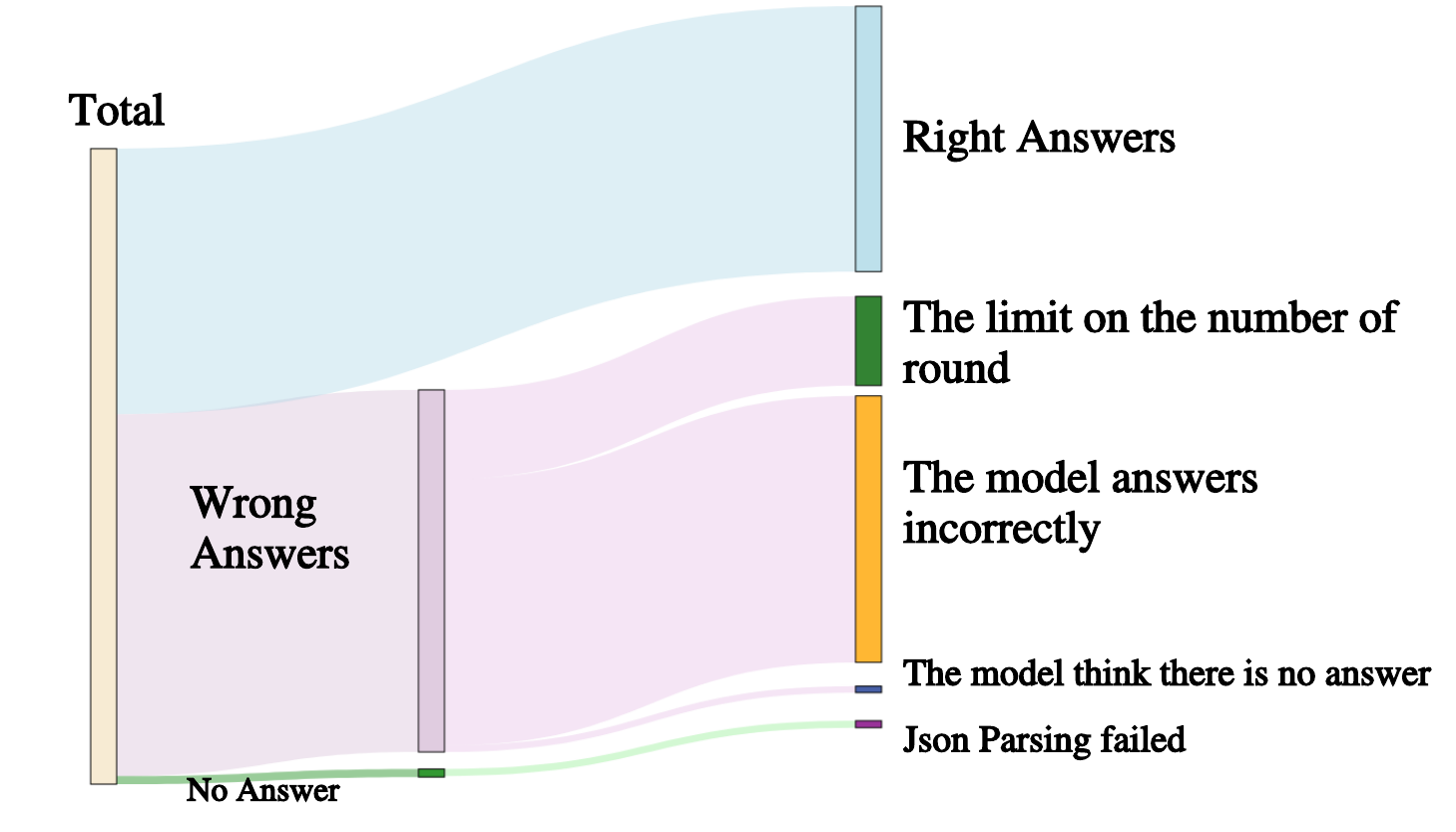}
  \captionof{figure}{Error analysis of MDTeamGPT on MedQA using LLaVA-1.6-7b as the model backend.}
  \label{Figure:error}
\end{wrapfigure}

We evaluate MDTeamGPT with the base model 
LLaVA-v1.6-mistral-7b-hf~\cite{liu2024improved} on the MedQA~\cite{jin2021disease} benchmark. We categorize the error types and plot them in Figure~\ref{Figure:error}. The failure samples account for 58.2\% of all samples. Among them, 41.9\% are due to incorrect model responses, and 14.0\% resulte from the failure caused by the round limit.For each type of error, we provide specific examples in Appendix D for analysis. This indicates that the majority of the performance loss comes from the model's reasoning itself and system-level constraints such as insufficient optimization of multi-round conversation strategies, which prevents the agent from converging to the correct answer within the limited interactions. Future optimizations should mainly focus on improving the alignment of the model's medical knowledge, reasoning, understanding capabilities, and dialogue management strategies.

%% file: Experiment_Details.tex
\section{Dataset}
\label{sec:Dataset}
This section details the categorization, preprocessing, and sample structure of the integrated clinical datasets. Table~\ref{tab:benchmark_datasets} summarizes the specifications of these datasets.

\noindent\textbf{Medical Question Answering}
\begin{itemize}[label=\textbullet]
    \item \textbf{MedQA:} MedQA is a medical text question-answer dataset that adopts a multiple-choice format. Its questions are drawn from examinations conducted by the United States, Mainland China, and Taiwan Medical Boards. These exams aim to assess a physician’s professional knowledge and clinical decision-making abilities. Our experiments primarily utilize the United States subset of this dataset~\cite{jin2021disease}.\\
   \noindent\textbf{Sample Question:} \textit{A 68-year-old male comes to the physician for evaluation of right flank pain. He has a history of diabetes and peripheral artery disease. His blood pressure is 160/90 mm Hg. Physical examination shows abdominal tenderness and right flank tenderness. An ultrasound shows dilation of the right ureter and renal pelvis. Which of the following is the most likely underlying cause of this patient's condition?}\\
   \noindent\textbf{Options:} \textit{A: Renal artery stenosis, B: Benign prostatic hyperplasia, C: Diabetic nephropathy, D: Common iliac artery aneurysm, E: Urethral stricture}
   
   \item \textbf{MedBullets:} MedBullets provides structured, crowd-sourced clinical knowledge derived from biomedical content to support rapid reference and decision-making in everyday practice~\cite{chen2025benchmarking}.\\
   \noindent\textbf{Sample Question:} \textit{A 42-year-old woman is enrolled in a randomized controlled trial to study cardiac function in the setting of several different drugs. She is started on verapamil and instructed to exercise at 50\% of her VO2 max while several cardiac parameters are being measured. During this experiment, which of the following represents the relative conduction speed through the heart from fastest to slowest?}\\
   \noindent\textbf{Options:} \textit{A: Atria > Purkinje fibers > ventricles > AV node, B: AV node > ventricles > atria > Purkinje fibers, C: Purkinje fibers > ventricles > atria > AV node, D: Purkinje fibers > atria > ventricles > AV node.}

    \item \textbf{MMLU:} MMLU Clinical Topics comprises real-world, multiple-choice questions across anatomy, clinical knowledge, professional medicine, medical genetics, college medicine and college biology. Each item centers on patient-centered scenarios to test practical clinical reasoning and guideline interpretation, emphasizing generalizability beyond memorization~\cite{hendrycks2020measuring}.\\
   \noindent\textbf{Sample Question:} \textit{A "dished face" profile is often associated with?}\\
   \noindent\textbf{Options:} \textit{A: a protruding mandible due to reactivation of the condylar cartilage by acromegaly., B: a recessive maxilla due to failure of elongation of the cranial base., C: an enlarged frontal bone due to hydrocephaly., D: defective development of the maxillary air sinus.}
   
    \item \textbf{VQA-RAD:} VQA-RAD (Visual Question Answering in Radiology) is a highly refined dataset with rich dimensions. The images are sourced from MedPix, and for each case, VQA-RAD selects only one representative image to ensure a unique patient for every image in the dataset~\cite{lau2018dataset}.\\
   \noindent\textbf{Sample Question:} \textit{Is there evidence of an aortic aneurysm?}\\
   \noindent\textbf{Image Path:} \textit{synpic42202.jpg}
\end{itemize}

\noindent\textbf{Medical Literature Reasoning}
\begin{itemize}[label=\textbullet]
   \item \textbf{PubMedQA:} PubMedQA is a biomedical question-answering (QA) dataset collected from PubMed abstracts. The task of PubMedQA is to answer research questions using the corresponding abstracts. We combine the original questions of PubMedQA with relevant background information to form a new question as the input for the model~\cite{jin2019pubmedqa}.\\
   \noindent\textbf{Sample Question:} \textit{Sternal fracture in growing children: A rare and often overlooked fracture? The relevant background information for this issue is as follows: Sternal fractures in childhood are rare. The aim of the study was to investigate the accident mechanism, the detection of radiological and sonographical criteria and consideration of associated injuries. In the period from January 2010 to December 2012 all inpatients and outpatients with sternal fractures were recorded according to the documentation. A total of 4 children aged 5-14 years with a sternal fracture were treated in 2 years, 2 children were hospitalized for pain management and 2 remained in outpatient care. The medical subject headings related to this issue include: Adolescent, Chest Pain, Child, Child, Preschool, Diagnosis, Differential, Fractures, Bone, Humans, Male, Rare Diseases, Sternum. In order to obtain an accurate answer, you may need complex reasoning.}\\
   \noindent\textbf{Options:} \textit{A: yes, B: maybe, C: no}
\end{itemize}

\noindent\textbf{Medical Visual Understanding and Reasoning}
\begin{itemize}[label=\textbullet]
   \item \textbf{SLAKE-En:} SLAKE is a bilingual English-Chinese dataset designed for medical visual question answering, comprising 642 images and 14,028 question-answer pairs. The dataset not only promotes automated interpretation of medical images but also enhances machine understanding and reasoning about medical image content through a QA format. We only evaluate the English portion of the test set from this dataset~\cite{liu2021slake}.\\
   \noindent\textbf{Sample Question:} \textit{What modality is used to take this image?}\\
   \noindent\textbf{Image Path:} \textit{xmlab102/source.jpg}
   
   \item \textbf{MedVidQA:} MedVidQA is a large medical teaching video QA dataset covering 899 YouTube medical teaching videos, totaling 95.71 hours. It yields 3,010 medical questions with corresponding answers. We augment the test set with Claude Sonnet4.5 to generate multiple-choice options (one correct, three distractors). To boost efficiency and maintain accuracy, we sample video frames, using 4 to 8 frames per video~\cite{gupta2023dataset}.\\
   \noindent\textbf{Sample Question:} \textit{How to perform chin tucks to treat neck pain?}\\
   \noindent\textbf{Options:} \textit{A: Lie supine with a rolled towel under the neck and lift the head off the surface, B: Lie supine with a rolled towel under the neck and gently push the back of the head into the surface, C: Lie supine with a rolled towel under the neck and rotate the chin toward the shoulder, D: Sit upright against a wall and push the back of the head into the wall.}\\
   \noindent\textbf{Video Path:} \textit{2711.mp4}
   
   \item \textbf{Med-CMR:} Med-CMR is a fine-grained Medical Complex Multimodal Reasoning benchmark containing seven tasks including causal reasoning, temporal prediction, fine-detail discrimination, multi-source integration, spatial understanding, small-object detection, and long-tail generalization, and each task corresponds to a specific type of medical multimodal reasoning complexity~\cite{gong2025med}.\\
   \noindent\textbf{Sample Question:} \textit{Based on the observed staining intensity and distribution in the provided image, which staining technique is most likely being used to highlight the abnormal cellular features in a rare disease setting?}\\
   \noindent\textbf{Options:} \textit{A: CK5 immunohistochemical staining, B: CK 6 immunostaining, C: CK5/6 immunofluorescence staining, D: CK 5/6 immunostaining, E: CK5/6 RNA in situ hybridization staining}\\
   \noindent\textbf{Image Path:} \textit{Med-CMR-mcq-3.png}
   
   \item \textbf{MedXpertQA-MM:} MedXpertQA-MM includes 4,460 questions across 17 specialties and 11 body systems, with two subsets: text and multimodal. We evaluate the multimodal subset (MedXpertQA MM)~\cite{zuo2025medxpertqa}.\\
   \noindent\textbf{Sample Question:} \textit{This patient presents for mammographic needle localization of a bar clip. What is the MOST optimal approach for needle localization?}\\
   \noindent\textbf{Options:} \textit{A: Lateral, B: Inferior, C: Medial, D: Superior, E: Oblique}
\end{itemize}

\noindent\textbf{Diagnosis Decision}
\begin{itemize}[label=\textbullet]
  \item \textbf{DxBench:} DxBench is a real-world automatic diagnosis evaluation dataset that is constructed based on real doctor-patient dialogues from MedDialog, structured and refined through extraction by GPT-4, and then manually verified. It includes over 1,000 real cases, covering 461 disease types from 15 departments and 5,038 symptoms~\cite{chen2025cod}.\\
   \noindent\textbf{Sample Question:} \textit{Here is a patient. Here are the details about this patient. He should go to the Surgery department. Among these data, the first item represents his physical characteristics, and the second item indicates whether the feature is true or not. Explicit symptoms: {Vomiting: True, Abdominal pain: True}. Implicit symptoms: {Yellow Sputum: True}. Which disease is he most likely to have?}\\
   \noindent\textbf{Options:} \textit{A: Cholangitis, B: Indigestion, C: Acute Enteritis}
\end{itemize}
   
\noindent\textbf{Evaluation of Medical Reasoning Chains}
\begin{itemize}[label=\textbullet]
   \item \textbf{M3CoTBench:} M3CoTBench is specifically designed to evaluate the correctness, efficiency, impact, and consistency of CoT reasoning in medical image understanding~\cite{jiang2026m3cotbench}.\\
   \noindent\textbf{Sample Question:} \textit{What is the most likely diagnosis?}\\
   \noindent\textbf{Options:} \textit{A: Age-related macular degeneration, B: Diabetic retinopathy, C: Conjunctivitis, D: Cataract}\\
   \noindent\textbf{Image Path:} \textit{4.png}
\end{itemize}

\begin{table}[h]
\vspace{-1em}
  \caption{Summary of the Medical Datasets. T: Text, I: Image, V: Video.}
  \label{tab:benchmark_datasets}
  \centering
  \scriptsize
  \renewcommand{\arraystretch}{1.05}
  \begin{tabular*}{\textwidth}{@{\extracolsep{\fill}}lcccl@{}}
    \toprule
    Dataset & Modality & Choice & Testing Size & Domain \\
    \midrule
    MedQA & T & MCQ & 1273 & US Medical Licensing Examination \\
    PubMedQA & T & MCQ & 500 & Medical Academic Question Answering \\
    MedBullets & T & MCQ & 308 & Online Platform Resources for Medical Study \\
    MMLU & T & MCQ & 1089 & Graduate Record Examination and US Medical Licensing Examination \\
    SLAKE-En & T, I & MCQ+Open-ended & 1061 & VQA Pairs Including CT, MRI and X-ray \\
    VQA-RAD & T, I & MCQ+Open-ended & 451 & VQA Pairs in Radiology \\
    MedVidQA & T, V & MCQ & 148 & First Aids, Medical Emergency, and Medical Education Questions \\
    Med-CMR & T, I & MCQ+Open-ended & 20654 & VQA pairs spanning 11 organ systems and 12 imaging modalities \\
    DxBench & T & MCQ & 1148 & A real-world diagnostic dataset with 1,148 real cases \\
    MedXpertQA-MM & T, I & MCQ & 2000 & VQA pairs spanning 17 specialties and 11 body systems \\
    M3CoTBench & T, I & MCQ+MRQ+Open-ended & 1078 & VQA dataset spanning 24 imaging modalities \\
    \bottomrule
  \end{tabular*}
\end{table}

%% file: MAS/methods.tex
\section{Methods}
Before detailing the specific adaptations for each method, Figure~\ref{Figure:10MAS} illustrates the simplified topological structures and interaction mechanisms of the representative multi-agent frameworks integrated into our platform.
\subsection*{Single Agent}
\noindent

\textbf{Chain-of-Thought (CoT)}~\cite{wei2022chain}. This baseline method tackles complex cognitive challenges by guiding models to generate a series of intermediate reasoning steps. Within our multimodal benchmark, we upgrade its standard prompting paradigm into a vision-driven medical chain-of-thought, compelling the single agent to explicitly cite key anatomical changes or lesion features within continuous video frames at every pathological deduction step, thereby significantly reducing hallucination rates during the zero-shot generation of medical evaluation data.

\textbf{Vision-Language Model (VLM)}~\cite{bai2025qwen3}. Represented by advanced foundational multimodal large models with official code at \href{https://github.com/QwenLM/Qwen3-VL}{GitHub repository}, these systems possess native capabilities for long-context and video temporal understanding. Serving as the core single-agent baseline in our benchmark, we specifically constraine its system instructions to meet stringent medical requirements, enabling it to directly parse extended medical video streams and generate strictly JSON-formatted correct clinical answers in an end-to-end manner alongside highly plausible distractors.

\subsection*{Multi-Agent Systems for General Tasks}
\noindent

\textbf{Self-Consistency}~\cite{wang2022self}. This method is originally designed to enhance the chain-of-thought performance of large language models by sampling multiple reasoning paths and executing a marginalization strategy to replace greedy decoding. To adapt it for complex medical video question-answering tasks, we extend its uni-modal decoding mechanism to construct a joint evaluation path that integrates medical semantic deduction with spatial-temporal visual features, ensuring the model achieves cross-modal logical self-consistency and visual alignment when outputting final structured clinical conclusions.

\textbf{AutoGen}~\cite{wu2024autogen}. The official code is located at \href{https://github.com/microsoft/autogen}{GitHub repository}. During our re-implementation, we entirely decouple its original general-purpose code execution logic, constructing a customized, heterogeneous topology comprising clinical experts and visual analysts exclusively for medical video VQA. We also enforce strictly standardized parsing modules to ensure the models conduct cross-disciplinary medical dialogues grounded entirely in multimodal evidence.

\textbf{Debate}~\cite{du2024improving}. The official code in \href{https://github.com/composable-models/llm_debate}{GitHub repository} implements multi-agent adversarial discussions. We transition its generic text-based prompts to a multimodal grounding protocol and replace its string-matching extraction with an VLM-assisted aggregator for robust clinical consensus generation.

\textbf{DyLAN}~\cite{liu2024dynamic}. The official repository is available at \href{https://github.com/SALT-NLP/DyLAN}{GitHub repository}. To adapt its dynamically evolving network to the medical domain, we completely overhaule its dynamic node evaluation scoring module, establishing clinical diagnostic accuracy and multimodal visual feature alignment as the sole criteria for node retention, effectively executing dynamic early-pruning of inferior nodes lacking medical justification.

\textbf{Discussion}~\cite{lu2024llm}. The official codebase in \href{https://github.com/lawraa/LLM-Discussion}{GitHub repository} facilitates decentralized multi-turn dialogue. We transform its free-form chat into a strictly governed clinical consultation board with a centralized adjudicator, mandating explicit visual evidence citations to reach a deterministic consensus.

\textbf{MetaPrompting}~\cite{suzgun2024meta}. The official repository in \href{https://github.com/suzgunmirac/meta-prompting}{GitHub repository} utilizes a centralized meta-agent to orchestrate specialized experts. We re-architect its routing heuristics to decompose tasks into parallel multimodal streams, implementing a weighted clinical-reconciliation protocol to synthesize visual and semantic medical data.

\textbf{ReConcile}~\cite{chen2024reconcile}. The official code in \href{https://github.com/dinobby/ReConcile}{GitHub repository} employs a structured reconciliation phase for conflict resolution. We profoundly extend this protocol into a multimodal grounding mechanism, forcing agents to trace logical disagreements back to specific visual timestamps using a confidence-weighted consensus algorithm.

\subsection*{Multi-Agent Systems for Medical Domains}
\noindent

\textbf{ColaCare}~\cite{wang2025colacare}. The official code in \href{https://github.com/PKU-AICare/ColaCare}{GitHub repository} models Electronic Health Records through multidisciplinary team (MDT) collaboration. We transform its text-centric ingestion pipeline to process sequential video frames, algorithmically forcing specialist agents to anchor diagnostic hypotheses to spatial-temporal anatomical anomalies.

\textbf{MDAgents}~\cite{kim2024mdagents}. The official code in \href{https://github.com/mitmedialab/MDAgents}{GitHub repository} provides an adaptive collaboration topology dynamically determined by medical query complexity. We re-architect its routing logic to concurrently evaluate multimodal spatial-temporal complexity, replacing its free-text output with a schema-driven aggregation module.

\textbf{MDTeamGPT}~\cite{chen2025mdteamgpt}. The official code in \href{https://github.com/KaiChenNJ/MDTeamGPT}{GitHub repository} simulates MDT consultations for static medical cases. We overhaul its case-presentation module to systematically process temporal video frames and impose strict multimodal grounding rules on inter-agent peer-review protocols to guarantee standardized evaluations.

\textbf{MedAgents}~\cite{tang2024medagents}. The official code in \href{https://github.com/gersteinlab/MedAgents}{GitHub repository} leverages a zero-shot, role-playing collaborative workflow. To integrate it into our multimodal benchmark, we inject vision-language bridging prompts during the independent analysis phase and introduce rigorous visual-verification constraints during cross-disciplinary debates to mitigate ungrounded medical hallucinations.

\textbf{MedAgentAudit}~\cite{gu2025medagentaudit}. This work originally aims to audit and quantify collaborative failure modes in medical multi-agent systems. We creatively transform this auditing mechanism into a real-time multimodal consensus monitoring node that actively detects and intervenes against spurious consensuses lacking visual evidence during the inference phase, effectively suppressing the masking of correct medical opinions and substantially improving QA generation quality.

\textbf{MedLA}~\cite{ma2025medla}. The official code is located at \href{https://github.com/alexander2618/MedLA}{GitHub repository}. This framework utilizes syllogistic logic trees to orchestrate complex medical reasoning. We anchor the foundational nodes of its logical deduction trees directly to key feature frames within the medical videos, ensuring that any deep-level medical inference is not only semantically self-consistent but also attains absolute evidence-based support in its visual morphology.

\textbf{CXRAgent}~\cite{lou2025cxragent}. The official code is located at \href{https://github.com/laojiahuo2003/CXRAgent/}{GitHub repository}. This framework originally employs a multi-stage reasoning pipeline orchestrated by a ``Director'' agent to analyze static chest X-rays. For our benchmark, we profoundly retrofit its ``Evidence-Driven Validator (EDV)'' with multimodal capabilities, enabling it to continuously evaluate dynamic temporal segments in medical videos. This ensures that all intermediate diagnostic plans are strictly anchored to the video's visual evidence before generating the final question-answer-distractor pairs.

\textbf{LINS}~\cite{wang2025lins}. The official code is available at \href{https://github.com/WangSheng21s/LINS}{GitHub repository}. This work introduces a multi-agent retrieval-augmented framework designed to generate citation-grounded medical texts. We refactor its Multi-Agent Iterative Retrieval-Augmented Generation (MAIRAG) algorithm to support the invocation of multimodal knowledge bases. This explicitly forces the agents to cross-validate with dynamic video frames when processing clinical queries, thereby yielding high-confidence and hallucination-free medical evaluation results.

\textbf{MedOrch}~\cite{he2025medorch}. This framework proposes providing flexible decision support by orchestrating domain-specific medical tools and reasoning agents. To seamlessly migrate it to our multimodal video VQA pipeline, we integrate an exclusive suite of spatiotemporal visual analysis tools. This compels the reasoning agents to continuously invoke and parse video-level evidence when generating complex clinical answers and their accompanying distractors, ensuring high traceability of the diagnostic process and deterministic structured outputs.

\textbf{MoMA}~\cite{gao2025moma}. This Mixture of Multimodal Agents architecture is initially designed to leverage multiple specialized large language models to fuse and convert multimodal electronic health records (EHR) into structured predictions. We thoroughly reshap its non-text modality transformation module, assigning its underlying "Specialist Agents" specifically to extract temporal pathological features from medical videos. This guarantees that the "Aggregator Agent" forms its final clinical consensus entirely based on a continuous vision-language alignment mechanism.

\begin{figure}[!h]
    \centering
    \includegraphics[width=0.9\textwidth]{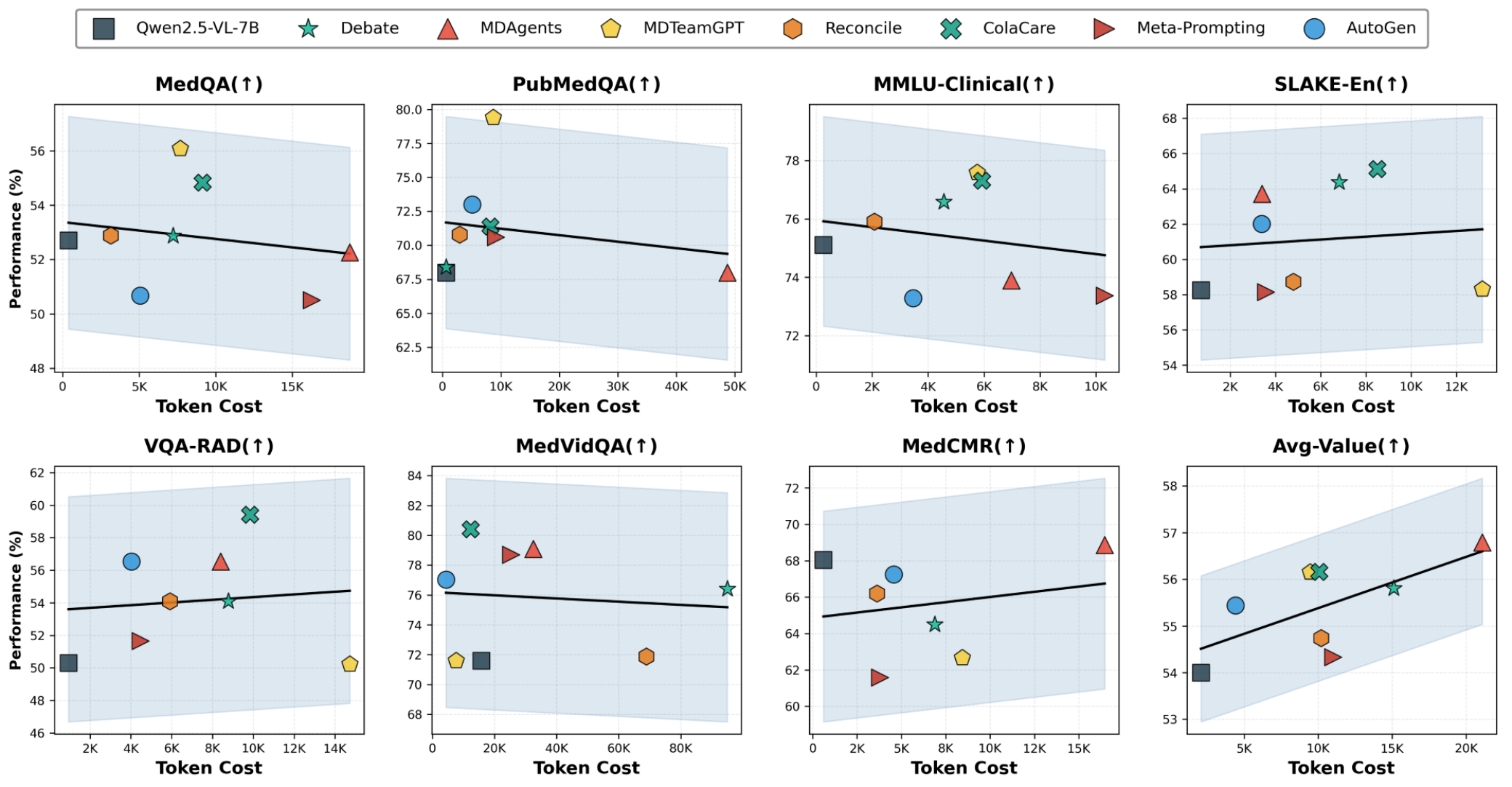}
    \caption{Trade-off between performance and token cost of Qwen2.5VL-7B-Instruct based Multi-Agent Methods across remaining Medical Benchmarks. We compute the average performance  and average token cost  across 11 datasets.}
    \label{fig:performance_vs_tokencost_qwen_7b_merged}
\end{figure}

\vspace{-1em}
\section{Additional Results}
\label{sec:Additional Results}
This section presents further experimental results and analysis to additionally support our conclusions. Our supplementary experimental findings and conclusions are presented in Figures 11, 12, and 13.

\begin{figure}[h]
    \centering
    \includegraphics[width=\linewidth]{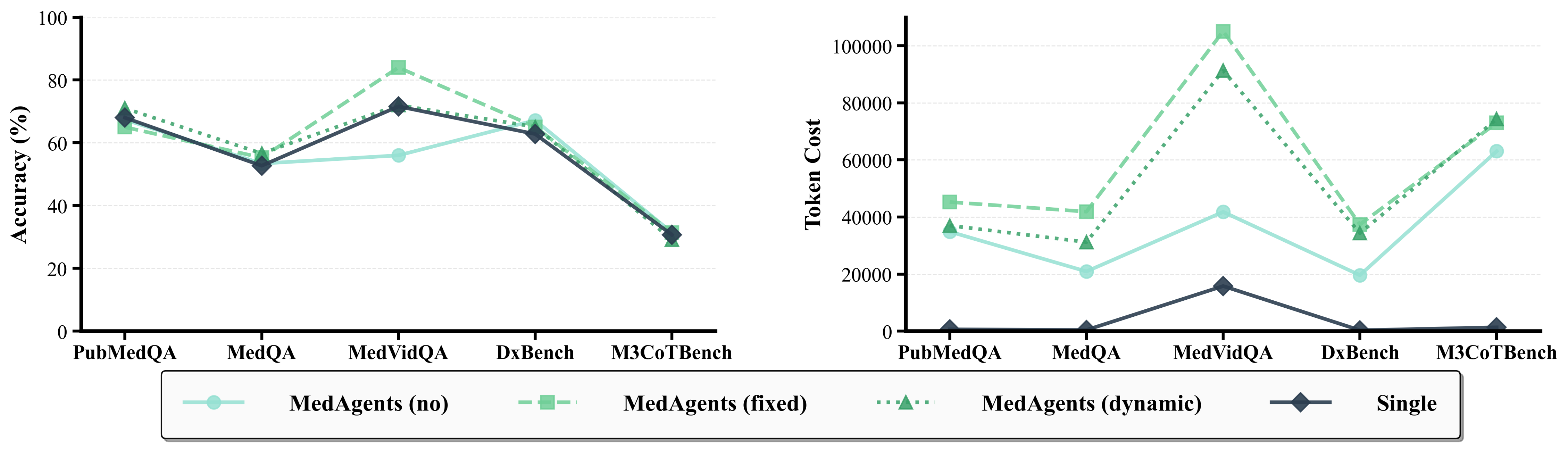}
    \caption{Performance and Token Cost of MedAgents under Different Medical Expert-Playing Modes.We use Qwen2.5VL-7B as the base model for MedAgents to investigate the performance and token cost of MAS frameworks on the PubMedQA, MedQA, MedVidQA, DxBench, and M3CoTBench datasets under three modes: no medical expert-playing, fixed medical expert-playing, and dynamic medical expert-playing.}
    \label{Figure:MedAgents_role_playing}
\end{figure}

\begin{figure}[!h]
    \centering
    \includegraphics[width=\linewidth, height=12cm]{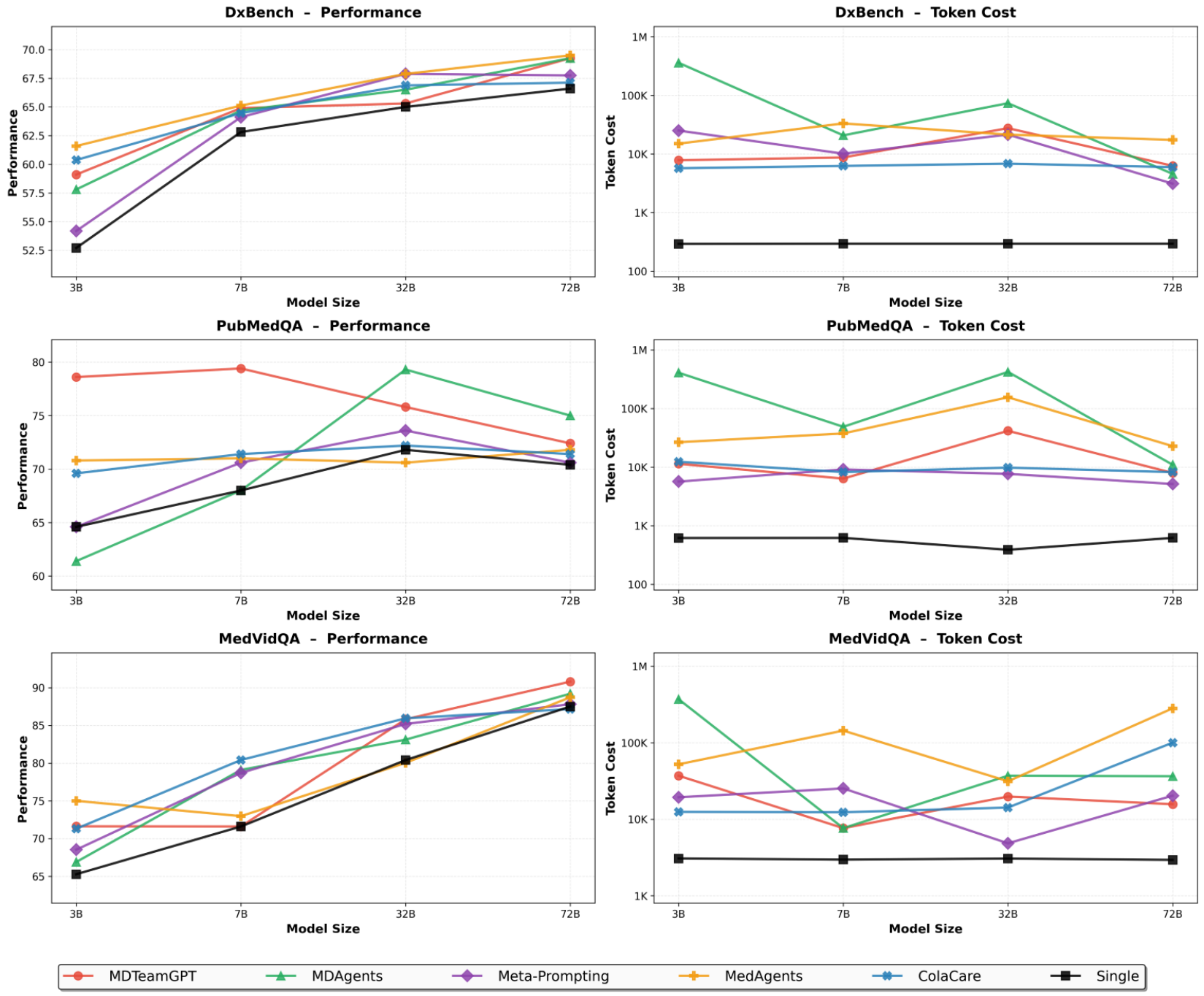}
    \caption{Backbone Model Size Scaling of Multi Agent Frameworks on PubMedQA, DxBench and MedVidQA with Performance and Token Cost Comparisons. Beyond our previous findings, we discover that increasing model scale does not necessarily lead to performance improvements for both multi-agent frameworks and single-agent systems. For instance, on the PubMedQA dataset at the 72B model scale, the performance of all MAS methods and single-agent models (except for the MedAgents framework) is lower than their 32B counterparts. Notably, MAS demonstrates an amplification effect on performance variations observed in single-agent models . For example ,improvements in single-agent performance are magnified in MAS, while performance degradation is similarly amplified.}
    \label{Figure:model_size_4_model}
\end{figure}

%% file: error.tex
\section{Error examples}
\label{sec:Error examples}
The error types we have are the following: The model answers incorrectly, The limit on the number of rounds, The model thinks there is no answer, JSON parsing failed, Each one is accompanied by a reference example.

\begin{tcolorbox}[
    colback=white,
    colframe=black,
    boxrule=3pt,              
    arc=5pt,                  
    left=15pt,
    right=15pt,
    top=4pt,                 
    bottom=4pt,
    title={JSON parsing failed},   
    titlerule=0pt,            
    coltitle=white,           
    colbacktitle=black        
]

\textbf{Question:} A 26-year-old male presents to his primary care physician with complaints of burning with urination, penile discharge, and intermittent fevers. A urethral smear shows gram negative diplococci within white blood cells. The organism grows well when cultured on Thayer-Martin agar. The patient is prescribed a course of ceftriaxone and the infection resolves without further complication. One year later, the patient returns with the same infection. Which of the following best explains this lack of lasting immunity?

\textbf{Options:}
\textbf{A)} Exotoxin release
\textbf{B)} Antigenic variation
\textbf{C)} Polysaccharide capsule
\textbf{D)} Bruton's agammaglobulinemia
\textbf{E)} Lack of necessary vaccination\\
\textbf{Answer:} Antigenic variation\\
\textbf{Model Answer:} \\
\textbf{Judge result:} Wrong
\end{tcolorbox}

\begin{tcolorbox}[
    colback=white,
    colframe=black,
    boxrule=3pt,              
    arc=5pt,                  
    left=15pt,
    right=15pt,
    top=4pt,                 
    bottom=4pt,
    title={The model answers incorrectly},   
    titlerule=0pt,            
    coltitle=white,           
    colbacktitle=black        
]

\textbf{Question:} A 1900-g (4-lb 3-oz) newborn is delivered at term to a 36-year-old primigravid woman. Pregnancy was complicated by polyhydramnios. Apgar scores are 7 and 7 at 1 and 5 minutes, respectively. He is at the 2nd percentile for head circumference and 15th percentile for length. Examination shows a prominent posterior part of the head. The ears are low-set and the jaw is small and retracted. The fists are clenched, with overlapping second and third fingers. The calcaneal bones are prominent and the plantar surface of the foot shows a convex deformity. Abdominal examination shows an omphalocele. Further evaluation of this patient is most likely to show which of the following findings?

\textbf{Options:}
\textbf{A)} Cataracts
\textbf{B)} Ventricular septal defect
\textbf{C)} Ebstein's anomaly
\textbf{D)} Pheochromocytoma
\textbf{E)} Holoprosencephaly\\
\textbf{Answer:} Ventricular septal defect\\
\textbf{Model Answer:} The most likely diagnosis for this newborn is pfeiffer syndrome.\\
\textbf{Judge result:} Wrong
\end{tcolorbox}

\begin{tcolorbox}[
    colback=white,
    colframe=black,
    boxrule=3pt,              
    arc=5pt,                  
    left=15pt,
    right=15pt,
    top=5pt,                 
    bottom=5pt,
    title={The limit on the number of rounds},   
    titlerule=0pt,            
    coltitle=white,           
    colbacktitle=black        
]

\textbf{Question:} An 11-month-old boy is brought to the emergency department by his mother after she observed jerking movements of his arms and legs for about 30 seconds earlier that morning. He has not had fever, cough, or a runny nose. He has been healthy, except for occasional eczema. He was delivered at home in Romania. His mother had no prenatal care. She reports that he has required more time to reach developmental milestones compared to his older brother. The patient's immunization records are not available. He takes no medications. He appears pale with blue eyes and has a musty odor. He has poor eye contact. Which of the following would have most likely prevented the patient's symptoms?

\textbf{Options:}
\textbf{A)} Levothyroxine therapy during pregnancy
\textbf{B)} Dietary restriction of phenylalanine
\textbf{C)} Daily allopurinol intake
\textbf{D)} Avoidance of fasting states
\textbf{E)} High doses of vitamin B6\\
\textbf{Answer:} Dietary restriction of phenylalanine\\
\textbf{Model Answer:} Max rounds reached. Proceeding with latest hypothesis.\\
\textbf{Judge result:} Wrong
\end{tcolorbox}

\begin{tcolorbox}[
    colback=white,
    colframe=black,
    boxrule=3pt,              
    arc=5pt,                  
    left=15pt,
    right=15pt,
    top=4pt,                 
    bottom=4pt,
    title={The model thinks there is no answer},   
    titlerule=0pt,            
    coltitle=white,           
    colbacktitle=black        
]

\textbf{Question:} A 48-year-old woman comes to the physician because of a 7-day history of progressively worsening shortness of breath. She has a history of asthma and has frequent exacerbations despite receiving maximal medical therapy. Physical examination shows scattered expiratory wheezing. Laboratory studies show a leukocyte count of 9,800/mm3 (13\% eosinophils) and elevated serum concentration of immunoglobulin E. An x-ray of the chest shows thickened bronchial walls and perihilar opacities. Sputum culture shows scattered septate hyphae that branch at 45 degrees. Which of the following is most likely to be increased in this patient?

\textbf{Options:}
\textbf{A)} IL-28
\textbf{B)} C5a
\textbf{C)} TGF-$\beta$
\textbf{D)} IFN-$\alpha$
\textbf{E)} IL-4\\
\textbf{Answer:} IL-4\\
\textbf{Model Answer:} None of the above.\\
\textbf{Judge result:} Wrong
\end{tcolorbox}

%% file: MAS/gui.tex
\section{MedMASLab Interactive Graphical User Interface}
\label{app:gui}

MedMASLab provides an interactive web-based graphical user interface to deliver accessible and standardized tooling for clinical research. This frontend replaces standard command-line execution with a visual environment, allowing researchers and clinicians to configure, customize, and evaluate multi-agent workflows without advanced programming expertise. The interface comprises five fundamental modules tailored to different phases of the evaluation lifecycle.

\begin{figure}[htbp]
    \centering
    \includegraphics[width=\textwidth]{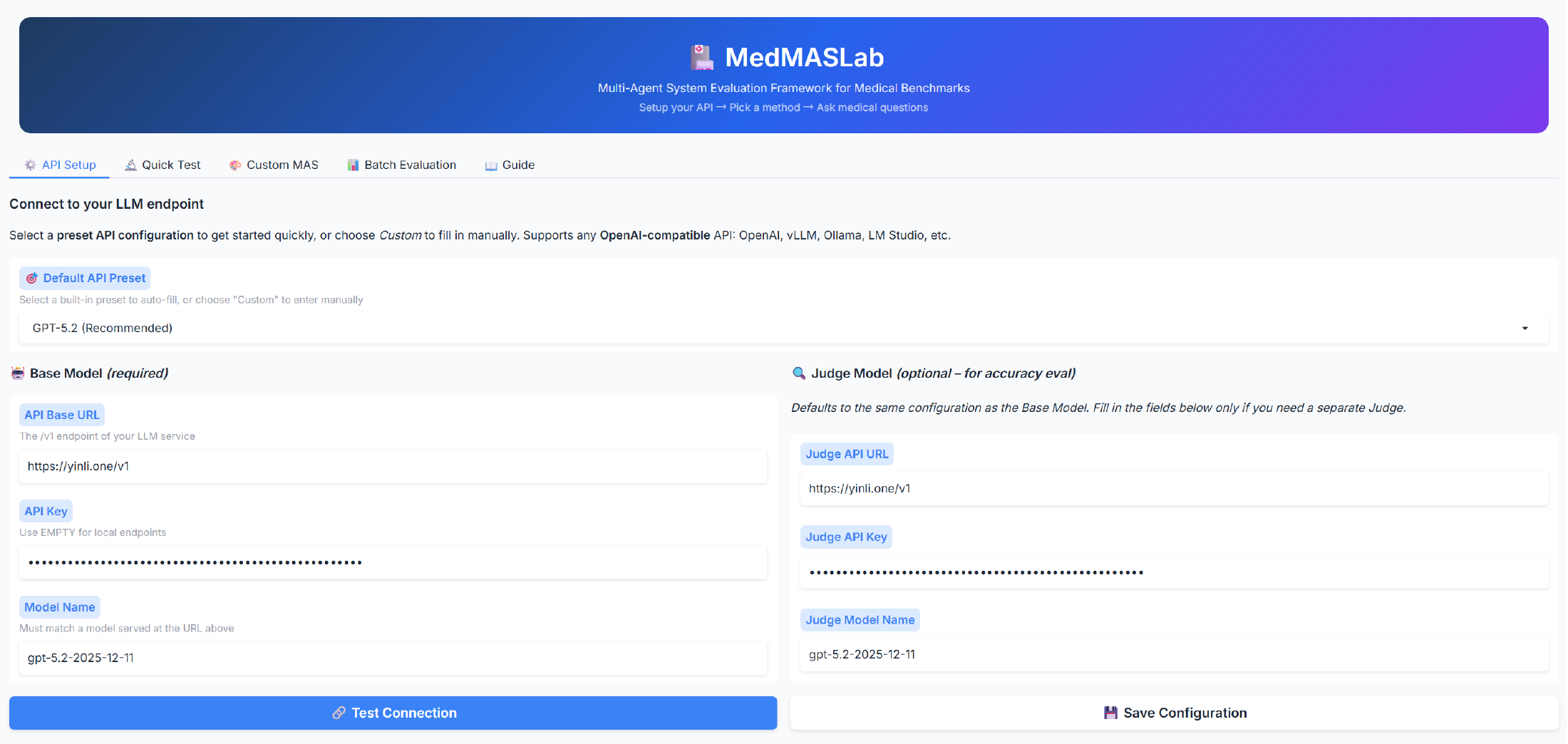}
    \caption{The API Setup interface, allowing decoupled configuration of Base and Judge models via OpenAI-compatible endpoints.}
    \label{fig:gui_setup}
\end{figure}

\noindent
\textbf{1. API Setup and Configuration.}
As illustrated in Figure~\ref{fig:gui_setup}, the \textbf{API Setup} interface acts as the central configuration hub for all computational resources. It supports connections to any OpenAI-compatible endpoint, including local deployments like vLLM and proprietary commercial services. This module enforces the structural decoupling of the framework by separating the inputs for the base reasoning model from the semantic evaluation judge. Furthermore, a built-in diagnostic tool verifies endpoint connectivity before initiating any inference tasks.

\begin{figure}[htbp]
    \centering
    \includegraphics[width=\textwidth]{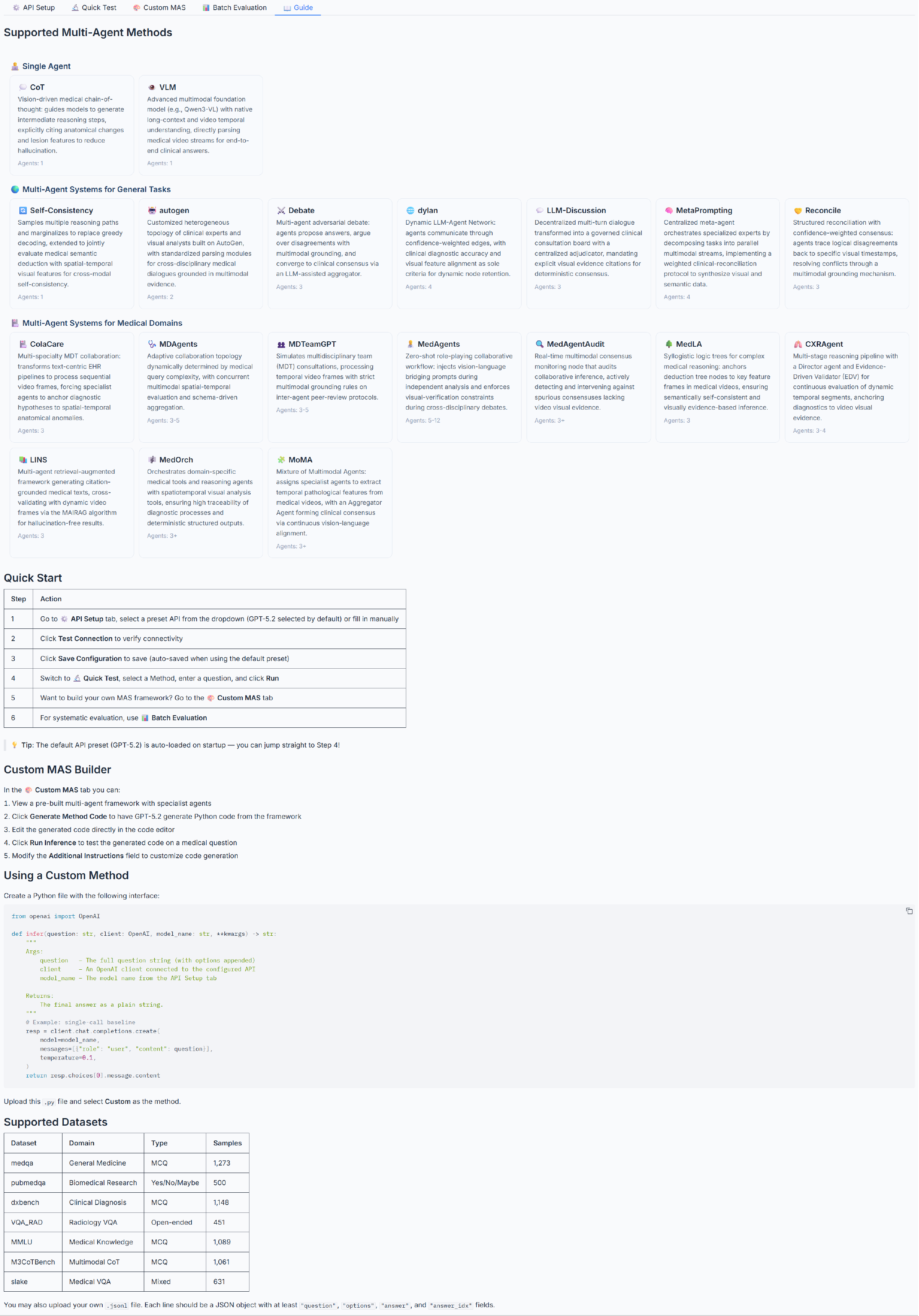}
    \caption{The Guide module, offering comprehensive documentation on supported MAS methods, dataset schemas, and tutorials.}
    \label{fig:gui_guide}
\end{figure}

\noindent
\textbf{2. Interactive Guide and Documentation.}
The \textbf{Guide} module, shown in Figure~\ref{fig:gui_guide}, provides an integrated documentation repository to assist user onboarding. It presents a clear taxonomy of all supported multi-agent methods, divided into general-purpose and medical-specific categories, alongside their respective algorithmic descriptions and agent capacities. The section also includes detailed data schemas for the supported clinical benchmarks and step-by-step tutorials to facilitate experimental setup.

\begin{figure}[htbp]
    \centering
    \includegraphics[width=\textwidth]{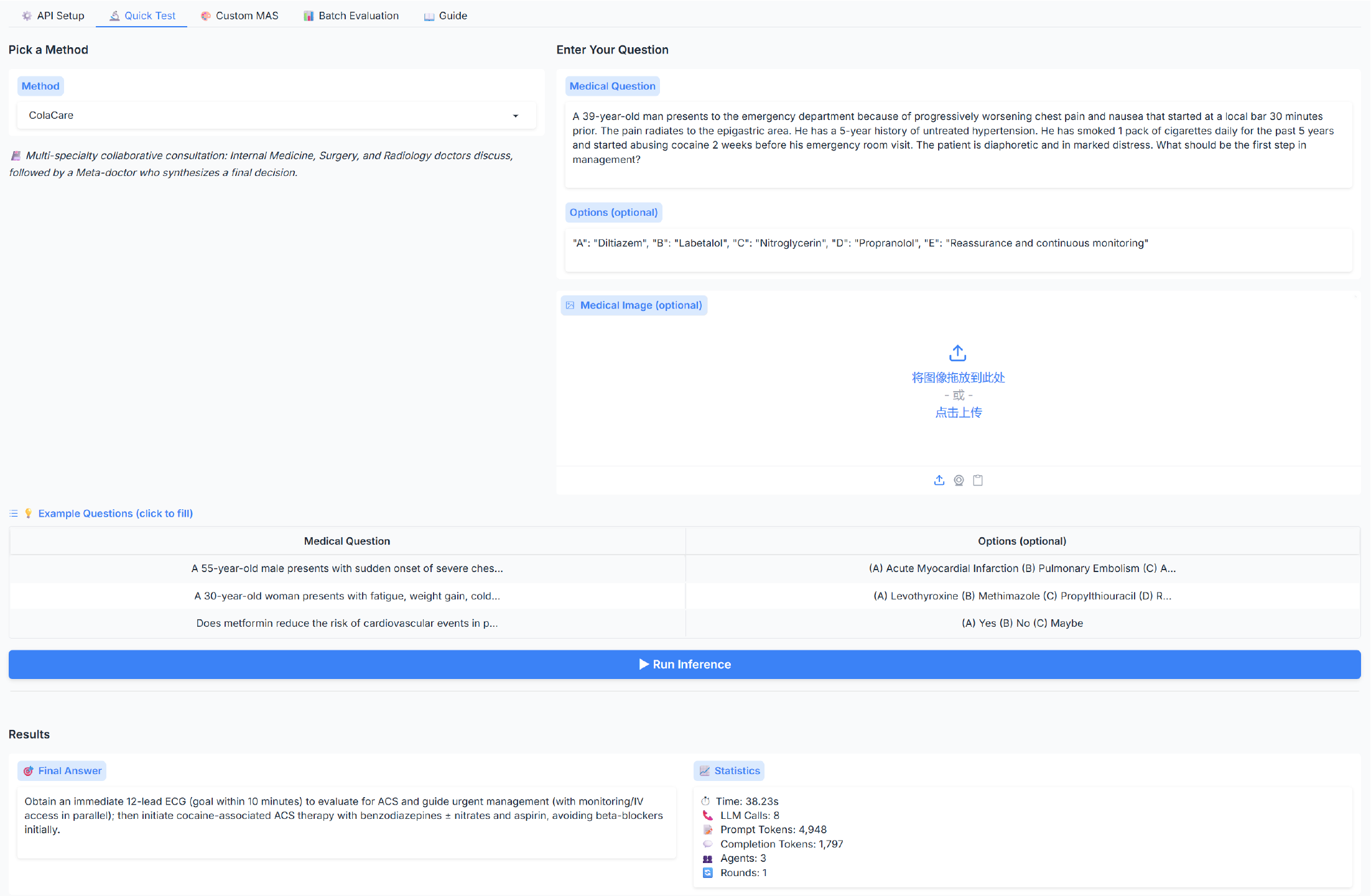}
    \caption{The Quick Test interface for single-sample multimodal inference and real-time performance profiling.}
    \label{fig:gui_quick_test}
\end{figure}

\noindent
\textbf{3. Single-Sample Quick Test.}
Figure~\ref{fig:gui_quick_test} presents the \textbf{Quick Test} module, which is designed for qualitative method analysis and rapid experimentation. Within this workspace, users select a pre-integrated baseline method and supply a clinical query along with optional supporting medical images. Following execution, the system displays the final diagnostic output alongside a detailed performance profile. This profile explicitly logs execution time, total VLM calls, token consumption, active agent counts, and the number of deliberation rounds.

\begin{figure}[htbp]
    \centering
    \includegraphics[width=\textwidth]{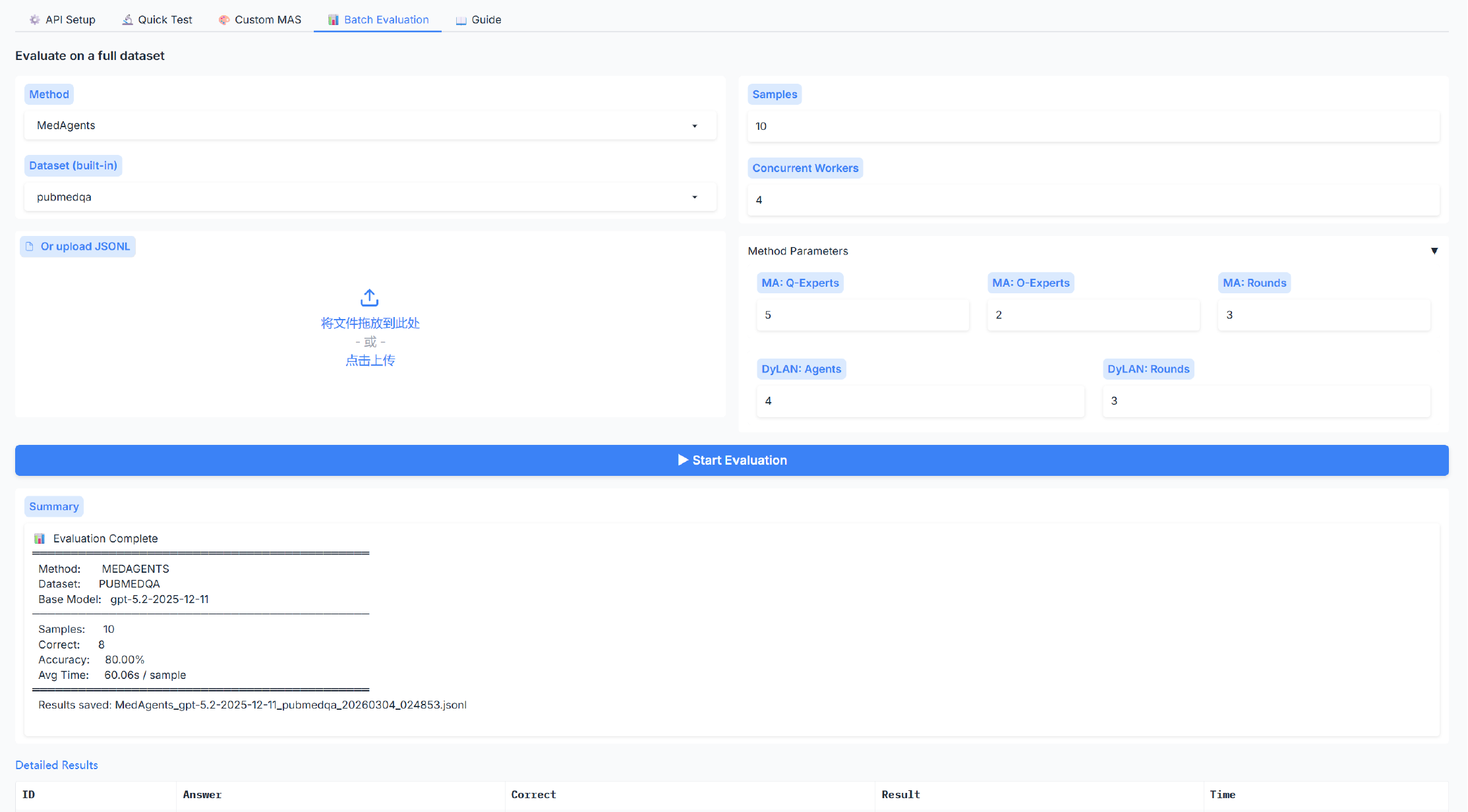}
    \caption{The Batch Evaluation module, orchestrating large-scale dataset testing with dynamic algorithmic parameter tuning.}
    \label{fig:gui_batch_eval}
\end{figure}

\noindent
\textbf{4. Large-Scale Batch Evaluation.}
The \textbf{Batch Evaluation} module detailed in Figure~\ref{fig:gui_batch_eval} facilitates large-scale benchmark campaigns. Users first select a target method and either a built-in dataset or a custom uploaded file. They then specify execution constraints, including maximum sample limits and concurrent worker threads. A key feature of this module is the dynamic generation of input fields corresponding to method-specific algorithmic parameters, such as the total count of expert agents or the maximum allowed debate rounds. This design enables systematic hyperparameter tuning and ablation studies, while a summary dashboard tracks accuracy and average latency, and supports data exportation.

\begin{figure}[htbp]
    \centering
    \includegraphics[width=\textwidth]{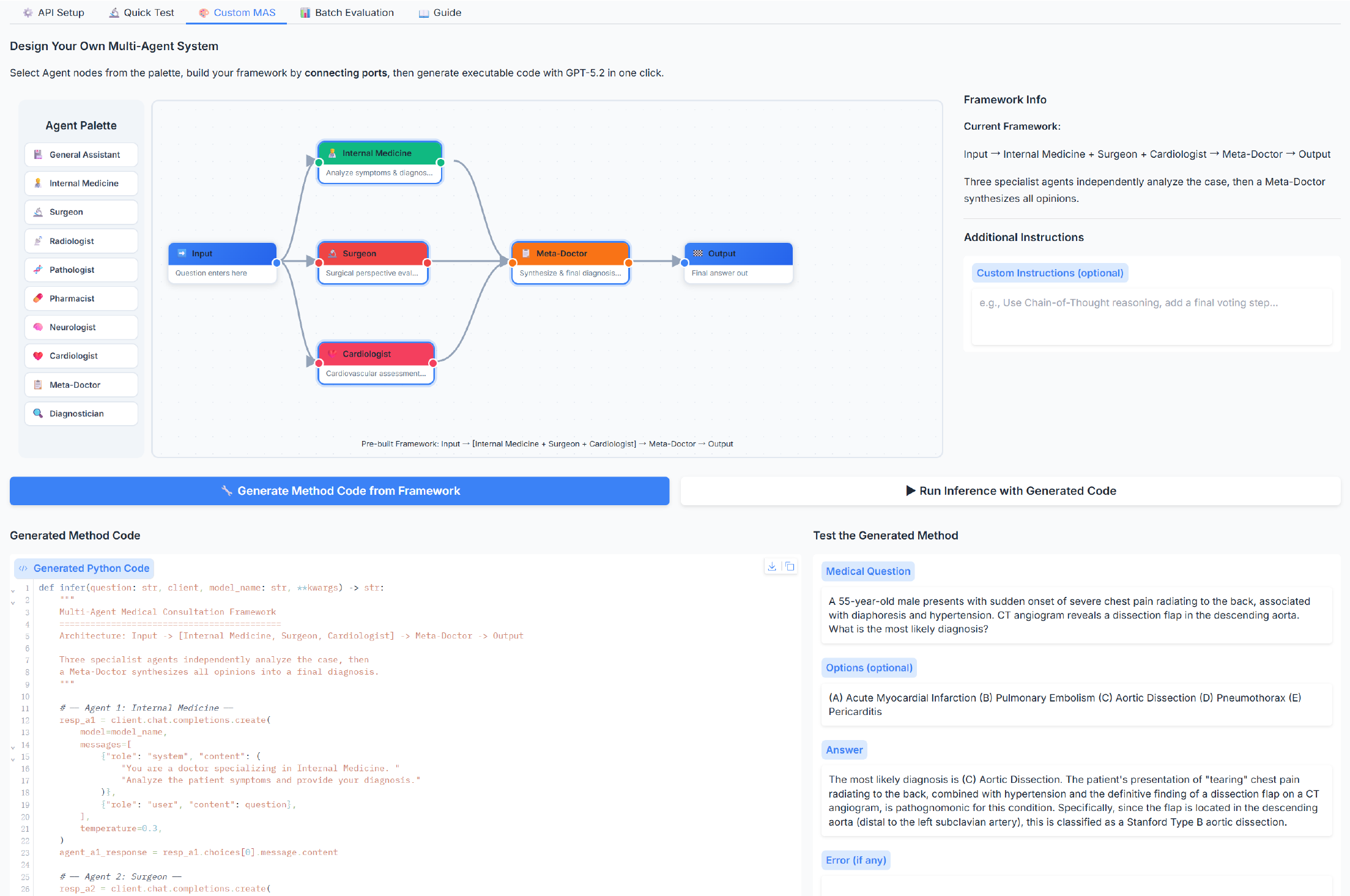}
    \caption{The Custom MAS Builder, providing a drag-and-drop canvas for topological design and automated Python code generation.}
    \label{fig:gui_custom_mas}
\end{figure}

\noindent
\textbf{5. Low-Code Custom MAS Builder.}
As depicted in Figure~\ref{fig:gui_custom_mas}, the \textbf{Custom MAS} module offers a visual, node-based canvas for architectural design. The interface provides an agent palette containing specialized medical roles like Surgeon, Radiologist, Pathologist, and Meta-Doctor. Users arrange these entities on the canvas and establish topological connections to define the desired communication workflow. An integrated generative model then translates this visual topology into executable Python code that strictly adheres to the unified inference standard of the MedMASLab framework. This automated generation allows researchers to immediately transition from conceptual topology design to empirical validation.